\documentclass{article}

\usepackage{multirow}
\usepackage{microtype}
\usepackage{graphicx}
\usepackage{subfigure}
\usepackage[table]{xcolor}  %

\usepackage[colorlinks=true, linkcolor=BrickRed, urlcolor=Blue, citecolor=Blue, anchorcolor=blue, backref=page]{hyperref}

\usepackage[accepted]{icml2025}

\usepackage{amsmath,amsfonts,bm}

\def\eqref#1{equation~\ref{#1}}

\def\1{\bm{1}}

\def\vb{{\bm{b}}}

\DeclareMathAlphabet{\mathsfit}{\encodingdefault}{\sfdefault}{m}{sl}
\SetMathAlphabet{\mathsfit}{bold}{\encodingdefault}{\sfdefault}{bx}{n}

\renewcommand*{\backref}[1]{}
\renewcommand*{\backrefalt}[4]{%
    \ifcase #1%
          \or [Cited on p.~#2.]%
          \else [Cited on p.~#2.]%
    \fi%
    }
\usepackage{url}
\usepackage[capitalize,noabbrev]{cleveref}
\numberwithin{equation}{section}
\newcommand{\mb}{{\mathbf{m}}}
\newcommand{\xb}{{\mathbf{x}}}
\newcommand{\Xb}{{\mathbf{X}}}
\newcommand{\Vb}{{\mathbf{V}}}
\renewcommand{\vb}{\mathbf{v}}
\newcommand{\Sigmab}{{\bm{\Sigma}}}
\newcommand{\Lambdab}{{\bm{\Lambda}}}

\usepackage{enumitem}  %
\usepackage{pifont} 
\usepackage{booktabs}
\usepackage{amsmath}
\usepackage{wrapfig}
\usepackage{amssymb}
\usepackage[symbol,hang,flushmargin]{footmisc}
\usepackage{adjustbox}

\newcommand{\norm}[1]{\left\lVert#1\right\rVert}

\newcommand{\rebuttal}[1]{\textcolor{black}{#1}}

\definecolor{lightgray}{gray}{0.8}
\definecolor{midgray}{gray}{0.5}
\definecolor{darkgray}{gray}{0.2}

\usepackage{amsmath}
\usepackage{amssymb}
\usepackage{mathtools}
\usepackage{amsthm}

\theoremstyle{plain}

\theoremstyle{definition}

\theoremstyle{remark}

\usepackage[textsize=tiny]{todonotes}

\icmltitlerunning{Eigenvector Masking for Visual Representation Learning}

\begin{document}

\twocolumn[
\icmltitle{From Pixels to Components:\\Eigenvector Masking for Visual Representation Learning}

\icmlsetsymbol{equal}{*}

\begin{icmlauthorlist}

\icmlauthor{Alice Bizeul$^{1,2}$}{}
\icmlauthor{Thomas M. Sutter$^{1}$}{}
\icmlauthor{Alain Ryser$^1$}{}
\icmlauthor{\textbf{Bernhard Schölkopf}$^2$}{}
\icmlauthor{\textbf{Julius von Kügelgen}$^3$}{} \icmlauthor{\textbf{Julia E. Vogt}$^1$}{}
\end{icmlauthorlist}

\icmlkeywords{Machine Learning, ICML}

\vskip 0.3in
]

\begin{abstract}
Predicting masked from visible parts of an image is a powerful self-supervised approach for visual representation learning. 
However, the common practice of masking random patches of pixels exhibits certain failure modes, which can prevent learning meaningful high-level features, as required for downstream tasks.
We propose an alternative masking strategy that operates on a suitable transformation of the data rather than on the raw pixels. 
Specifically, we perform principal component analysis and then randomly mask a subset of components, which accounts for a fixed ratio of the data variance.
The learning task then amounts to reconstructing the masked components from the visible ones.
Compared to local patches of pixels, the principal components of images carry more global information. 
We thus posit that predicting masked from visible components involves more high-level features, allowing our masking strategy to extract more useful representations.  
This is corroborated by our empirical findings which demonstrate improved image classification performance for component over pixel masking.
Our method thus constitutes a simple and robust data-driven alternative to traditional masked image modeling approaches\footnote[0]{Preprint. Under Review.}\footnote[2]{We release the code to reproduce our results on \href{https://github.com/alicebizeul/pmae}{GitHub}. Correspondence to \href{mailto:alice.bizeul@inf.ethz.ch}{alice.bizeul@inf.ethz.ch}}.\footnote[0]{$^1$ Department of Computer Science, ETH Zürich $^2$ MPI for Intelligent Systems, Tübingen $^3$ Seminar for Statistics, ETH Zürich}
\end{abstract}

\section{Introduction}

In masked image modeling~\citep[MIM;][]{pathak2016context}, parts of an image are masked, and a model has to reconstruct the missing parts from the visible ones---analogous to predicting hidden words in masked language modeling~\citep{devlin2018bert}.
To succeed at this task, it is thought that the model must learn a meaningful representation of the visual content~%
\citep{kong2023understanding}. 
Empirically, this approach indeed produces 
representations that perform well when fine-tuned on downstream tasks, such as image classification and semantic segmentation~\citep{zhou2021ibot,bao_beit_2022,xie2022simmim,baevski_data2vec_2022,dong2023peco}.

\begin{figure}
    \centering
    \vspace{-0.2em}
    \includegraphics[width=\linewidth]{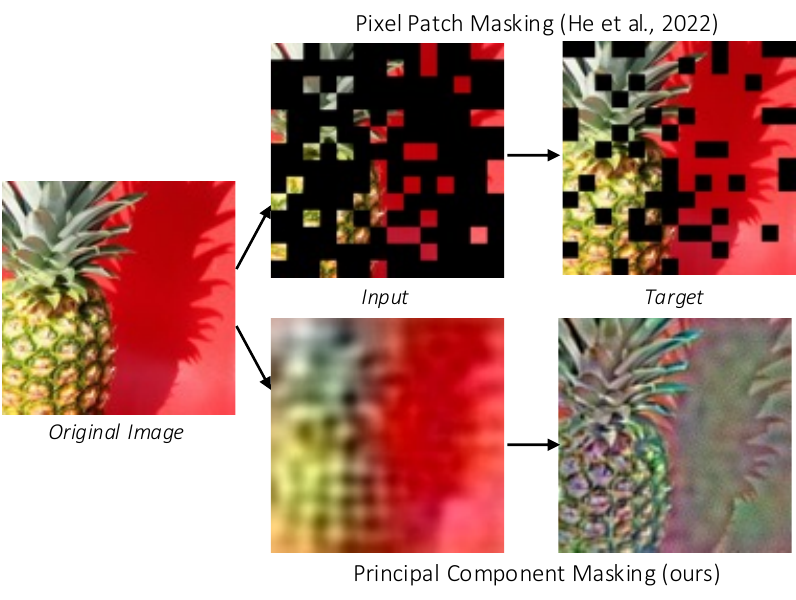}
    \vspace{-2.2em}
    \caption{\small \textbf{From Pixels to Components.} 
    Masked image modeling involves reconstructing masked-out patches of pixels from visible ones.
    Instead of masking in pixel space (top), we propose applying random masks to a transformed version of the image, specifically to its principal component representation (bottom). Two disjoint sets of components are used as input and reconstruction target.
    }
    \vspace{-1.3em}
    \label{fig:pitch}
\end{figure}

MIM has been particularly effective when combined with vision transformers~\citep[ViT;][]{dosovitskiy_image_2021}.
A prominent example is the masked autoencoder~\citep[MAE;][]{he_masked_2021}, consisting of
a ViT encoder-decoder architecture and a %
masking strategy that randomly selects a fixed ratio of square image patches, see~\cref{fig:pitch} (top). The encoder processes the visible patches and the decoder aims to reconstruct the masked content from the inferred representation.

\begin{figure*}[t]
    \vspace{-5pt}
    \centering
    \includegraphics[width=\textwidth]{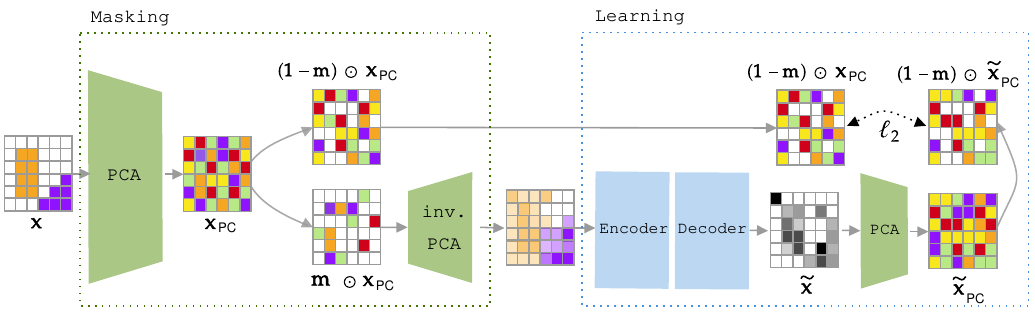}
    \vspace{-0.8cm}
    \caption{\small \textbf{Overview of the Principal Masked Autoencoder.} A principal masked  autoencoder (PMAE) differs from a vanilla MAE~\citep{he_masked_2021} by performing the masking in the space of principal components $\xb_\text{PC}\!=\!\text{PCA}(\xb)$ rather than in pixel space. The visible principal components $\bm \mb \odot \xb_{\text{PC}}$ are then projected back into the observation space and serve as the input for a ViT encoder-decoder architecture. Masked principal components, $(\mathbf{1}\!-\!\mb) \odot \xb_{\text{PC}}$, serve as the reconstruction target.}
    \vspace{-0.3cm}
    \label{fig:overview}
\end{figure*}

While the inner workings of MAEs remain poorly understood~\citep{zhang_how_2022,yue_understanding_2023}, \citet{kong2023understanding} suggested an explanation from a latent variable model perspective. By splitting an image into two separate views and asking the model to predict one from the other, MAEs are compelled to pick up any information that is shared across views. If the partition into views (i.e., the masking strategy) is chosen \textit{carefully}, this shared information will include high-level latent variables, such as object class. Since solving common downstream tasks with a simple (e.g., linear) predictor precisely requires identifying such high-level features, this offers a possible explanation for the observed effectiveness of MAE representations. 

\begin{figure}[t]
    \centering
    \includegraphics[width=0.48\textwidth]{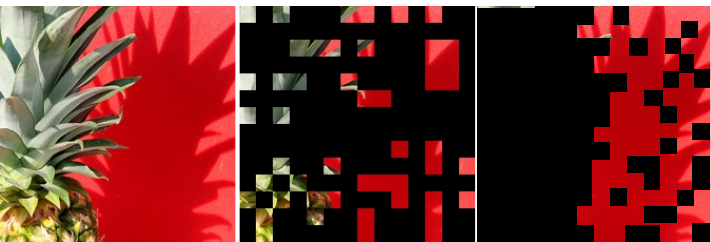}
    \vspace{-18pt}
    \caption{\small \looseness-1 \textbf{Masking in pixel space.} Image (left) with a random spatial mask \textit{partially} removing relevant information (middle) and removing \textit{all} semantic information (right). The latter is an example in which MIM would likely fail to learn useful representations.}
    \vspace{-10pt}
    \label{fig:pixel_mask}
 \end{figure}

\looseness-1
With this in mind, we
ask: \emph{Is masking random patches in pixel space the optimal strategy for MIM?} 
In natural language, each word in a sentence tends to carry semantic information, and the information shared between sets of words often conveys the general message of the sentence. 
However, this does not necessarily apply to the visual domain, where
individual pixels or entire patches commonly contain redundant information (e.g., background).
Moreover, (small) objects can be masked out completely %
such that any information about them is lost and 
reconstruction becomes 
impossible, see~\cref{fig:pixel_mask}~(right). 
\looseness-1 Some works have thus 
sought to devise more elaborate masking strategies which leverage auxiliary information such as learned or given image segmentations~\citep{li_mst_2021,kakogeorgiou_what_2022,shi_adversarial_2022}.
Without such prior knowledge or %
complex training pipelines to identify the structure of an image, randomly masking a fixed ratio of patches of pixels remains the default practice.  
Yet, relying on this masking strategy assumes---rather unrealistically---that the information shared between any random partition of pixel patches naturally aligns with high-level variables of interest~\citep{kong2023understanding}. 

\begin{figure*}[t]
    \centering
    \includegraphics[width=1.01\linewidth]{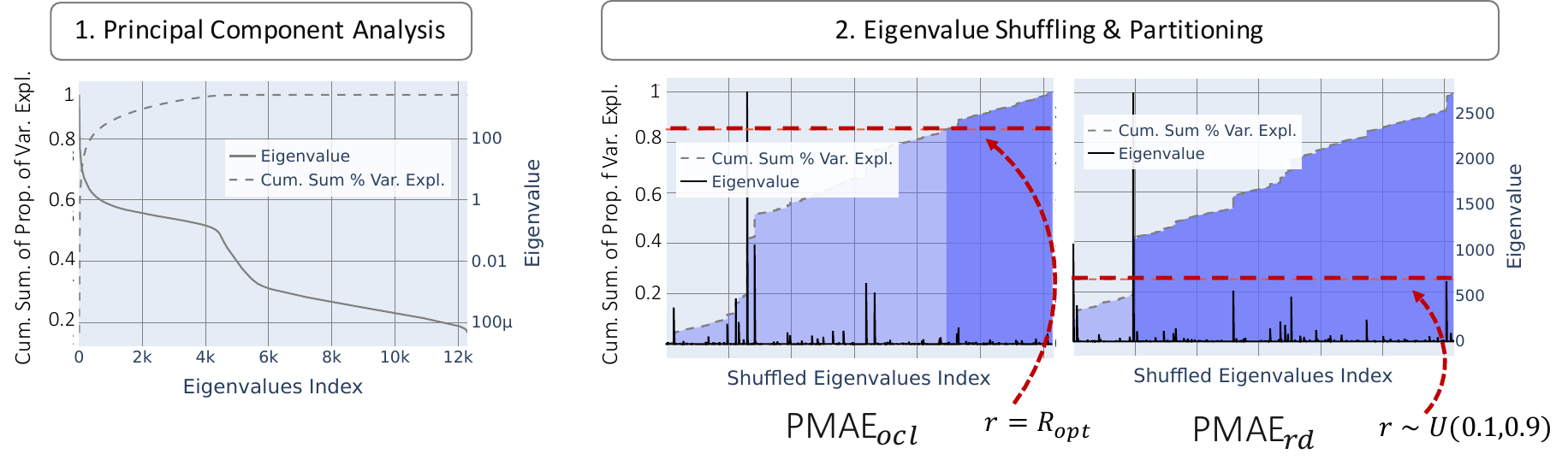}
    \vspace{-25pt}
    \caption{\small \looseness-1 \textbf{Mask Design in PMAE.} 1.\ Perform PCA 2.\ For each batch, randomly shuffle the principal components and select a subset to construct the input (light blue), while the remaining components are used to create the reconstruction target (dark blue). In PMAE$_{\text{ocl}}$, the input components are chosen to explain $100\!\times\!(1\!-\!r)\%$ of the data's variance with $r$ optimized for downstream performance (here $R_{\text{opt}}\!=\!0.15$). In PMAE$_{\text{rd}}$, the input explains between 10\% and 90\% of the variance with $r$ sampled independently and uniformly in $[0.1,0.9]$ for each batch.}
    \vspace{-0.3cm}
    \label{fig:masking3}
\end{figure*}

\looseness-1 In this work, we introduce an \textit{alternative data-driven masking strategy for MIM}.
Rather than working directly in pixel space, we propose to first project images into a latent space and then perform random masking on the transformed data. 
Among the infinitely many candidate transformations, we opt for simplicity and choose principal component analysis (PCA), a well-established linear pre-processing technique that is deterministic and (hyper)parameter-free.
Following the PCA transformation, some principal components are masked-out and need to be reconstructed from the remaining visible components, see~\cref{fig:pitch} (bottom) for an example.
Further, we leverage the fact that each principal component captures a known fraction of the data variance to inform our masking strategy. 
Specifically, we mask a random subset of principal components that accounts for a fixed ratio of the variance~(\cref{fig:masking3}). 
The ratio of masked variance serves as a proxy for the complexity of the modeling task making it a more interpretable and more easily tunable hyperparameter than the ratio of masked patches~(\cref{fig:mae_hyperparameters}, left).
We combine this masking strategy with the MAE architecture and refer to the resulting method as \textit{principal masked autoencoder} (PMAE), see~\cref{fig:overview} for an overview.

\looseness-1 
Relying on a transformation like PCA allows for partitioning the information in an image into a set of global features rather than into local patches of pixels, see \Cref{fig:pitch} . 
This can help overcome the aforementioned failure modes of spatial masking where the input and target share too much (redundancy) or too little (impossibility) information.
The space of principal components may, therefore, constitute a more meaningful domain for masking, resulting in more useful high-level representations.
 
This is consistent with recent work, highlighting the beneficial partitioning of image information by PCA. \citet{balestriero_learning_2024} demonstrate that low-eigenvalue components capture features crucial for common downstream tasks~(see also~\cref{fig:masking_pc}), and \citet{chen2024deconstructing} highlight the importance of the space in which image distortions are applied, referring to PCA as a valuable transformation to consider. To the best of our knowledge, our work is the first to leverage such insights to devise a simple, robust, and effective data-driven alternative to pixel-space masking in MIM.

In experiments on the CIFAR10, TinyImageNet and three medical MedMNIST datasets, we observe our approach (PMAE) to yield superior performance to the default strategy of spatial masking (MAE), while being less sensitive to the choice of masking ratio hyperparameter.
These empirical findings support our claim that masking principal components instead of pixel patches can facilitate the learning of more meaningful high-level representations.

\section{Background}\label{sec:mae}
\textbf{Principal Component Analysis.}

\looseness-1 
Principal component analysis \citep[PCA;][]{pearson1901,hotelling1933analysis} identifies components in data that exhibit the highest variance. Given a centered data matrix~$\Xb\in\mathbb{R}^{N\times D}$ with $N$ observations of dimension $D$, PCA seeks weight vectors $\vb_l\in\mathbb{R}^D$ for $l=1, \ldots, L\leq D$ (the principal components or PCs) that maximize the variance of the projections $\Xb\vb_l$, subject to orthogonality with previously found PCs and unit-length constraints. The solution to this problem is given by the eigenvalue decomposition of the empirical covariance matrix~$\Sigmab=\Xb^\top\Xb$, i.e., $\Sigmab=\Vb\Lambdab\Vb^\top$, where $\Lambdab=\mathrm{diag}(\lambda_1, \ldots, \lambda_D)$ contains the eigenvalues $\lambda_1>\ldots>\lambda_D$, and $\Vb \in \mathbb{R}^{D \times D}$ contains the corresponding eigenvectors. The first $L$ PCs correspond to the eigenvectors of $\Sigmab$ with the largest eigenvalues, capturing the dominant modes of variation in the data, with the variance explained by each PC  proportional to its eigenvalue~$\lambda_l$. While PCA is often used with $L<D$ for dimensionality reduction, we focus on the lossless case where $L=D$. The projection of $\Xb$ onto its principal components (\text{``into PC space"}) is given by $\Xb_\text{PC}=\Xb\Vb$, and the inverse transformation by $\Xb=\Xb_\text{PC}\Vb^\top$. 
By construction, PCA produces features which are uncorrelated ($\Xb_\text{PC}^\top\Xb_\text{PC}=\bm\Lambda$). However, we emphasize that---except for special cases such as when the data is multivariate Gaussian---this does \textit{not} imply independence. 
In general, the values of a subset of PCs are therefore (nonlinearly) predictive of the values of other PCs.
Our results in~\cref{sec:results} and examples of reconstructions of masked principal components in \Cref{app_sec:reconstruction} demonstrate that this is the case for the natural and medical images we consider.

\looseness-1
\textbf{Representation Learning.} Representation learning~\citep{bengio2013representation} aims at learning an embedding function or encoder $f:\mathbf{x}\mapsto\mathbf{z}$, which maps observations $\mathbf{x}\in\mathbb{R}^D$ to representations $\mathbf{z}\in\mathbb{R}^K$. These representations are meant to capture some of the explanatory factors underlying the data, thus making~them well-suited for use in downstream tasks such as predicting a target variable~$y$ (e.g., the class or location of objects), often thought of as simple functions of the high-level explanatory factors. 

\textbf{Masked Image Modeling.} Prominent approaches to representation learning in the image domain rely on the masked image modeling paradigm \citep{pathak2016context,zhou2021ibot,he_masked_2021,bao_beit_2022}. Here we focus on the widely adopted MAE \citep{he_masked_2021} as a representative of MIM. 
A MAE consists of: an encoder~$f_{\bm{\phi}}$, parametrized by~$\bm{\phi}$, which maps the visible portions of the input together with their positional embeddings to a representation; and a decoder $g_{\bm{\theta}}$, parametrized by~$\bm{\theta}$, which reconstructs the missing parts from their positional embeddings and the representation produced by the encoder.

Given an observation $\xb\in\mathbb{R}^D$ and  complementary binary masks $\mb, (\bm 1-\mb)\in\{0,1\}^D$, which extract the visible and masked parts, respectively, the MAE objective is given by
\begin{equation}
\begin{aligned}
\label{eq:mae}
    &\mathcal{L}_{\text{MAE}}(\mathbf{x}, \mathbf{m}; \bm\theta, \bm\phi)
    =
    \\
    &\norm{
    \left(\mathbf{1}-\mathbf{m}\right)\odot
    \Big[
    g_{\bm{\theta}}\circ f_{\bm{\phi}}\left(\mathbf{m} \odot \mathbf{x}\right)
    -\mathbf{x} 
    \Big]
    }_2^2,
    \end{aligned}
\end{equation}
where $\odot$ denotes element-wise multiplication and $\circ$ denotes function composition.
The parameters $(\bm\phi, \bm\theta)$ are optimized via stochastic gradient descent on~\cref{eq:mae}. 
 
\looseness-1 The mask~$\mathbf{m}$ partitions the $D$ pixels into two disjoint sets of $(1-r)D$ visible and $rD$ masked out pixels, where $r$ is referred to as the \textit{masking ratio}.
Typically, $\mb$ is chosen by partitioning the image into square patches (thus introducing patch size as an additional hyperparameter) and then randomly selecting a ratio of $r$ patches to be masked.
 
Prior work has relied on hyperparameter sweeps to identify the masking ratio and patch size that optimize downstream performance \citep{he_masked_2021,zhang_how_2022}. These efforts have led to the widely adopted approach of masking out $75\%$ ($r=0.75$)
of patches of size $16\times 16$ pixels.

\looseness-1
\textbf{Challenges Resulting from Spatial Masking.}
Even though  MIM with random spatial masking produces strong results on representation learning benchmarks~\citep{dong2023peco}, it is based on the rather unrealistic assumption that, given any partition of image patches into two disjoint sets, the information shared between views contains variables that are linearly predictive of downstream targets~$y$~\citep{kong2023understanding}. 
As visualized in~\Cref{fig:pixel_mask}, %
for some masks (such as the one in the middle) %
shared information indeed includes features such as object type.
However, for other masks (such as the one on the right)
predicting the correct class label %
from the visible patches is almost impossible. 
The latter therefore yields input-target pairs which do not contribute a useful self-supervised learning signal.
Moreover, even for well-designed masks, many masked-out and visible patches contain redundant information such as background. 
This leads us to conjecture that %
spatial masking is a suboptimal approach to MIM that is not always well-aligned with common downstream tasks and may suffer from slow convergence and high sensitivity to hyperparameters, as suggested by prior work \citep{he_masked_2021,balestriero_learning_2024} and confirmed in \cref{fig:mae_hyperparameters}. 

\section{Principal Masked Autoencoders}
To address the challenges arising from spatial masking, we propose an alternative masking strategy and a corresponding MAE-variant which we refer to as \textit{principal masked  autoencoder} (PMAE). PMAE builds on the MIM learning paradigm but differs from prior approaches by performing the masking operation on a latent space.
Specifically, we first apply an invertible transformation $t:\mathbb{R}^{D}\rightarrow\mathbb{R}^{D}$ to an image $\xb$ and then mask and reconstruct at the level of $t(\xb)$, 
resulting in the following objective:
\begin{equation}
\begin{aligned}
\label{eq:pmae_lossA}
    &\mathcal{L}_{\text{PMAE}}\left(\mathbf{x},\mathbf{m};\bm\theta,\bm\phi\right)= 
    \\ 
    &\norm{(\mathbf{1}-\mathbf{m})\odot \Big[ 
    t\circ g_{\bm{\theta}}\circ f_{\bm{\phi}}\circ t^{-1}\left(\mathbf{m}\odot t\left(\mathbf{x}\right)\right)
    - t(\mathbf{x})
    \Big]
   }_2^2
   . \nonumber
\end{aligned}
\end{equation}
\looseness-1 
Similar to \cref{eq:mae}, 
$f_{\bm{\phi}}$ is an embedding function which encodes visible parts of the image
and $g_{\bm{\theta}}$ is a decoder which reconstructs the missing parts. Note that \Cref{eq:mae} is recovered as a special case of \Cref{eq:pmae_lossA} if $t$ is chosen to be the identity mapping. 

\textbf{Design Choices.} \looseness-1 While \Cref{eq:pmae_lossA} can generally accommodate any invertible mapping $t$, this work specifically explores the use of PCA as a data transformation, i.e., we choose $t(\mathbf{x})=\mathbf{x}_\text{PC}=\xb\Vb$.
\Cref{fig:overview} provides a visual summary of PMAE based on PCA.
We leave an investigation of alternative image transformations (e.g., the Fourier transform) for future work, see~\Cref{sec:discussion} for further discussion.

\looseness-1 Following MAEs, we use ViTs \citep{dosovitskiy_image_2021}  for both $f_{\bm\phi}$ and $g_{\bm\theta}$. Since these architectures are optimized for processing images, 
we project the visible components back into image space using the inverse PCA transformation~$t^{-1}$ before encoding. Similarly, to reconstruct the missing components, the decoder output is projected back into PC space using~$t$.
ViTs are a common choice of architecture for MIM due to the use of binary masks in pixel space. These masks produce images with black-patch occlusions (see~\cref{fig:pixel_mask}), which are poorly aligned with the inductive biases of other architectures.
By masking principal components instead of patches of pixels, the encoder inputs and reconstruction targets in PMAEs display smoother activation patterns (see \Cref{fig:pitch}, bottom).
PMAEs are therefore, in principle, also compatible with, e.g., convolutional networks.  
An exploration of other architectural choices for PMAE is left for future work.

\begin{figure*}[t]
    \centering
    \vspace{-0.2cm}
    \begin{subfigure}
    \centering
    \begin{minipage}{0.7\textwidth}
    \centering
    \includegraphics[width=\textwidth]{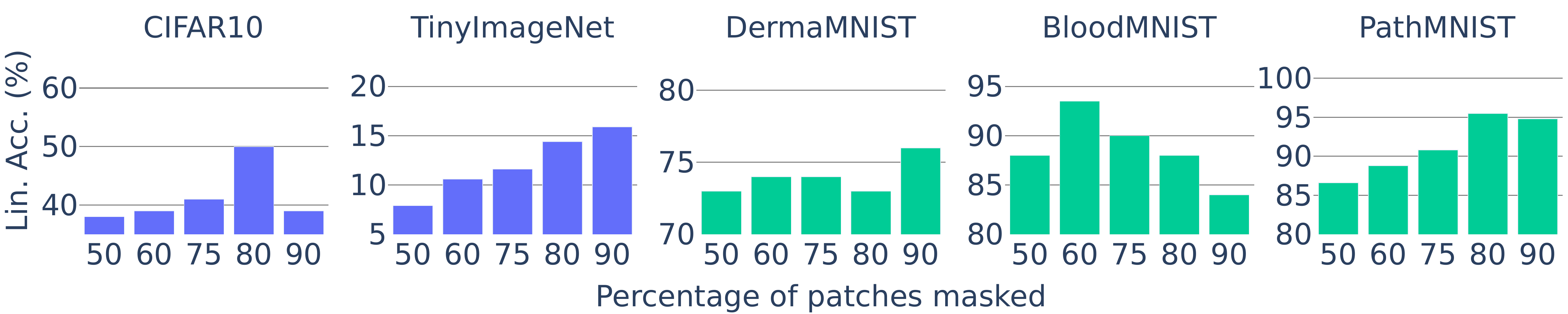} 
    \vspace{-0.2cm}
    \includegraphics[width=\textwidth]{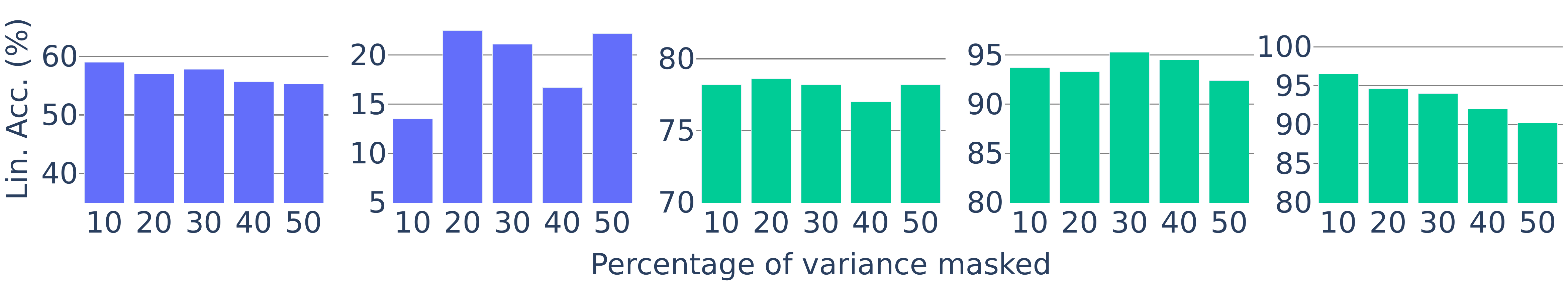} 
    \end{minipage}
    \end{subfigure}%
    \begin{subfigure}
    \centering
    \begin{minipage}{0.29\textwidth}
    \vspace{-5pt}
    \includegraphics[width=\textwidth]{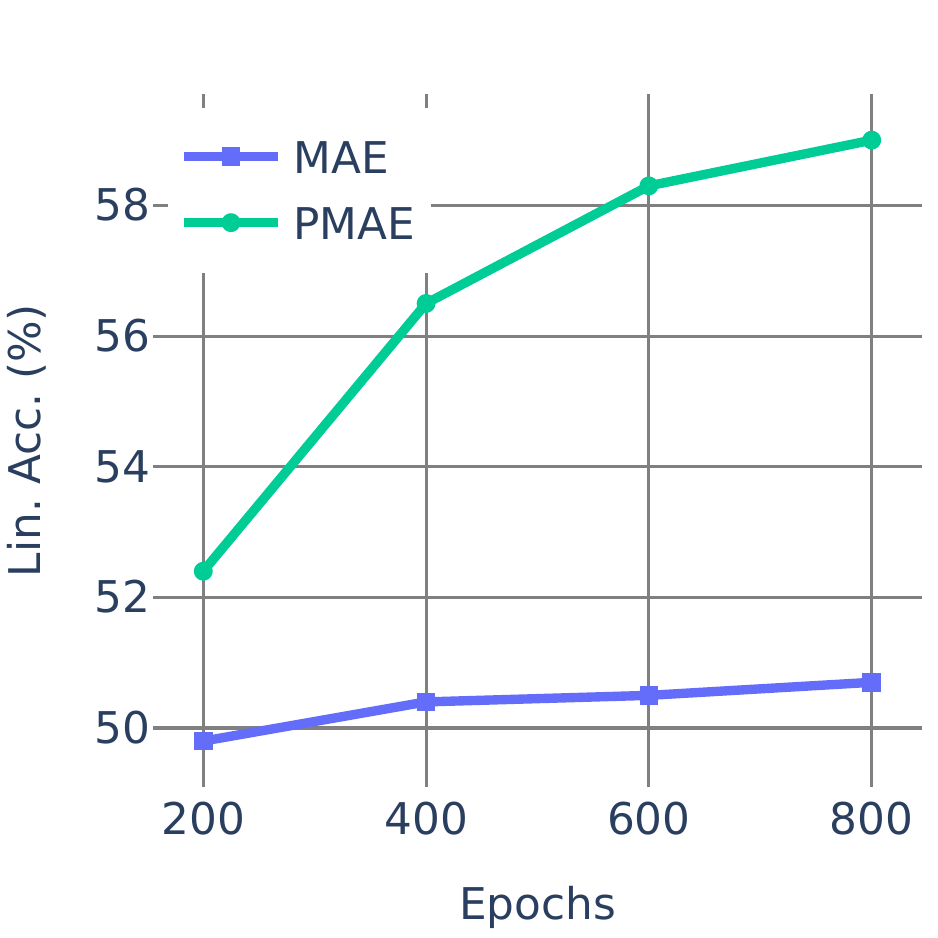} 
    \end{minipage}
    \end{subfigure}
    \vspace{-0.1cm}
    \caption{\small 
    \textbf{Impact of the Masking Ratio.} MAE (top) and PMAE (bottom) linear probing accuracy for varying masking ratios. The masking ratio is a sensitive and data-dependent hyper-parameter. While for MAE a clear masking guideline is hard to extract, for PMAE we observe close-to-optimal performance across datasets for $20\%$ of the data variance masked.  
    \textbf{(Right) Learning curves.} Linear probe accuracy for CIFAR10 classification across training epochs. PMAE outperforms MAE's final performance after 200 epochs.
    }
    \vspace{-0.3cm}
    \label{fig:mae_hyperparameters}
\end{figure*}

\looseness-1 
Analogous to the MAE objective in~\Cref{eq:mae}, the PMAE objective 
only penalizes reconstruction errors on the part masked out by $(\bm 1-\mb)$. In \Cref{eq:pmae_lossA}, the model output is projected into PC space, and the $l_2$ norm is computed between the predicted and ground truth  masked-out principal components. 
In \Cref{app:reconstruct_pc}, we explore an alternative objective for which the reconstruction error is calculated directly in the image space. 
Specifically, \Cref{eq:pmae} computes the $l_2$ norm between the decoder output and the masked-out principal components projected back into the image space. However, this alternative approach results in lower downstream performance, likely due to the stricter constraints placed on the decoder which, in this case, also needs to predict the effect of the visible components---akin to penalizing an MAE decoder for in-painting, rather than predicting perfectly black patches for the visible parts.

\looseness-1 Mask design is also a crucial aspect of our approach. In MAEs, a fixed percentage of pixel patches is masked out. The masking ratio is usually chosen based on a hyperparameter tuning conducted for each individual dataset (see \Cref{fig:mae_hyperparameters}, top). Instead in PMAE, we choose to systematically mask components that collectively account for a fixed percentage of the variance of the dataset. \Cref{fig:masking3} illustrates this process. After applying PCA, we randomly shuffle the order of principal components within each batch and divide them into two disjoint sets: one explaining \(100(1-r)\%\) of the variance, and the other, \(100r\%\), where \(r\) is our masking ratio. The ratio of masked variance serves as a more interpretable and easily tunable hyperparameter compared to the masked patch ratio used in MAEs (\Cref{fig:mae_hyperparameters}, bottom), effectively acting as a proxy for modeling task complexity. Further details on the choice of \(r\) are provided in \Cref{sec:setup}.

\textbf{Intuition behind PMAE.} \looseness-1 Contrary to spatial masking presented in \cref{sec:mae}, an appropriate function $t$, in \Cref{eq:pmae_lossA}, can encourage information that is shared between visible and masked-out information to contain $y$. More specifically, if the latent space captures unique global information in each dimension, masking any of these dimensions retains information about all parts of the image. Hence, \Cref{eq:pmae_lossA} allows us to learn more meaningful representations for a suitable choice of $t$. While there may be many appropriate choices for $t$, we found that applying PCA and projecting samples using the resulting principal components is a suitable choice for the latent space. In particular, each dimension captures specific factors of variation observed within the dataset and is typically tied to global features as shown in \Cref{fig:masking_pc}. Masking one factor of variation thus prevents us from completely removing all information about variables of interest within a sample, as most principal components will retain some information about them. 

\looseness-1 In PMAE, the masking ratio—the proportion of explained variance masked—can also easily be interpreted, as it directly corresponds to a well-understood quantity. In contrast, masking pixel patches lacks a straightforward connection to a well-defined metric. Depending on the dataset, a given proportion of masked patches can correspond to varying amounts of information content, making the masking ratio in the spatial domain harder to interpret.

\section{Experimental Setup}\label{sec:setup}
We now outline the setup used to validate \(\text{PMAE}\).  
Our experiments adapt the implementation proposed by \citet{he_masked_2021}.
Further experimental details and information about computational costs can be found in \Cref{app:setup}.

\textbf{Mask Design.} Following MAE \citep{he_masked_2021}, we take random image patch masking as our baseline. Based on ablation studies from \citet{he_masked_2021}, the standard practice involves masking out 75\% of image patches (denoted as MAE$_{\text{std}}$). We also examine an oracle-based masking strategy (denoted as MAE${_\text{ocl}}$), where the masking ratio is tuned to optimize linear probing downstream performance. This setting serves as an upper bound to the downstream performance. Additionally, we introduce a randomized masking approach, MAE${_\text{rd}}$, in which the masking ratio is independently sampled within a range of 0.1 to 0.9 for each batch. %
This strategy is exempt from any hyperparameter tuning and offers insights into the robustness of methods under suboptimal hyperparameters. 

A similar approach is applied to PMAE, where we consider both oracle (PMAE$_{\text{ocl}}$) and randomized (PMAE$_{\text{rd}}$) masking strategies. In the oracle approach, we define the optimal percentage of \textit{variance} to be masked based on linear probe downstream performance on a held-out dataset. In the randomized strategy, we ensure at least $10\%$ and at most $90\%$ of the data variance is masked out. This percentage is independently sampled for each batch. \Cref{fig:masking} provides examples of the images obtained from these masking strategies. %

\textbf{Training \& Evaluation.} \looseness-1 We train a ViT-Tiny~\citep{touvron2021training,dosovitskiy_image_2021} encoder and decoder backbones. Following \citet{he_masked_2021}, we use image flipping and random image cropping as data augmentations. The decoder's output is normalized \citep{he_masked_2021} for MAEs only as this did not result in a consistent performance gain for PMAE. We train representations for 800 epochs and provide an overview of the evolution of performance across training in \Cref{app:results}. We then evaluate learned representations on image classification using a linear probe and multi-layer perceptron (MLP) classifier on top of the encoder's output \texttt{[CLS]} token which is frozen. Additionally, we also explore the fine-tuning setting in which the pre-trained encoder and a linear probe appended to the \texttt{[CLS]} token are trained simultaneously. Following \citet{he_masked_2021}, we fix the training duration at evaluation to 100 epochs. \Cref{app:results} also reports downstream performance obtained with a $k$-NN classifier.\

\begin{table}[t]
\centering
\caption{\small\looseness-1 Linear \& MLP probe and fine-tuning top-$1\%$ accuracy for CIFAR10, TinyImageNet and MedMNIST datasets for random masking in pixel (MAE) and principal component (PMAE) space with the standard $75\%$ (std), oracle (ocl) and random (rd) masking ratios. $^\star$ refers to ours.}
\vspace{5pt}
\label{tab:results}
\resizebox{\columnwidth}{!}{ %
\begin{tabular}{clccccc}
\toprule
&& \small{\textbf{CIFAR10}}   & \small{\textbf{TinyIN}}  & \small{\textbf{Derma}}   & \small{\textbf{Blood}} & \small{\textbf{Path}}  \\
\midrule
\multirow{5}{*}{\textbf{Linear}} 
& \cellcolor{gray!25} MAE$_{\text{std}}$  & \cellcolor{gray!25} $41.7$  & \cellcolor{gray!25} $11.5$  & \cellcolor{gray!25} $72.4$  & \cellcolor{gray!25} $73.4$  & \cellcolor{gray!25} $83.4$  \\
& \cellcolor{gray!10} MAE$_{\text{ocl}}$  & \cellcolor{gray!10} $50.7$  & \cellcolor{gray!10} $15.5$  & \cellcolor{gray!10} $73.7$  & \cellcolor{gray!10} $78.6$  & \cellcolor{gray!10} $86.4$  \\
& \cellcolor{gray!10} PMAE$_{\text{ocl}}^{\star}$ & \cellcolor{gray!10} $\mathbf{59.0}$ & \cellcolor{gray!10} $\mathbf{22.5}$ & \cellcolor{gray!10} $\mathbf{78.6}$ & \cellcolor{gray!10} $\mathbf{95.5}$ & \cellcolor{gray!10} $\mathbf{96.8}$ \\
& MAE$_{\text{rd}}$   & $41.9$  & $7.5$   & $72.4$  & $83.2$  & $85.6$  \\
& \cellcolor{gray!0} PMAE$_{\text{rd}}^{\star}$ & \cellcolor{gray!0} $\mathbf{44.0}$ & \cellcolor{gray!0} $\mathbf{16.9}$ & \cellcolor{gray!0} $\mathbf{76.4}$ & \cellcolor{gray!0} $\mathbf{90.0}$ & \cellcolor{gray!0} $\mathbf{90.1}$ \\
\midrule
\multirow{5}{*}{\textbf{MLP}} 
& \cellcolor{gray!25} MAE$_{\text{std}}$  & \cellcolor{gray!25} $34.0$  & \cellcolor{gray!25} $15.5$  & \cellcolor{gray!25} $72.2$  & \cellcolor{gray!25} $68.6$  & \cellcolor{gray!25} $92.6$  \\
& \cellcolor{gray!10} MAE$_{\text{ocl}}$  & \cellcolor{gray!10} $55.2$  & \cellcolor{gray!10} $22.2$  & \cellcolor{gray!10} $74.4$  & \cellcolor{gray!10} $75.8$  & \cellcolor{gray!10} $95.1$  \\
& \cellcolor{gray!10} PMAE$_{\text{ocl}}^{\star}$ & \cellcolor{gray!10} $\mathbf{64.1}$ & \cellcolor{gray!10} $\mathbf{25.1}$ & \cellcolor{gray!10} $\mathbf{80.2}$ & \cellcolor{gray!10} $\mathbf{92.5}$ & \cellcolor{gray!10} $\mathbf{98.6}$ \\
& \cellcolor{gray!0} MAE$_{\text{rd}}$   & \cellcolor{gray!0} $38.5$  & \cellcolor{gray!0} $11.6$  & \cellcolor{gray!0} $66.9$  & \cellcolor{gray!0} $70.6$  & \cellcolor{gray!0} $95.7$  \\
& \cellcolor{gray!0} PMAE$_{\text{rd}}^{\star}$ & \cellcolor{gray!0} $\mathbf{47.0}$ & \cellcolor{gray!0} $\mathbf{22.6}$ & \cellcolor{gray!0} $\mathbf{77.4}$ & \cellcolor{gray!0} $\mathbf{80.2}$ & \cellcolor{gray!0} $\mathbf{97.7}$ \\
\midrule
\multirow{5}{*}{\shortstack[l]{\textbf{Fine-}\\ \textbf{tuned}}} 
& \cellcolor{gray!25} MAE$_{\text{std}}$  & \cellcolor{gray!25} $75.7$  & \cellcolor{gray!25} $37.5$  & \cellcolor{gray!25} $80.4$  & \cellcolor{gray!25} $97.8$  & \cellcolor{gray!25} $\bm{99.7}$  \\
& \cellcolor{gray!10} MAE$_{\text{ocl}}$  & \cellcolor{gray!10} $80.5$  & \cellcolor{gray!10} $42.8$  & \cellcolor{gray!10} $79.9$  & \cellcolor{gray!10} $\bm{98.1}$  & \cellcolor{gray!10} $\bm{99.7}$  \\
& \cellcolor{gray!10} PMAE$_{\text{ocl}}^{\star}$ & \cellcolor{gray!10} $\mathbf{84.8}$ & \cellcolor{gray!10} $\mathbf{44.5}$ & \cellcolor{gray!10} $\mathbf{82.3}$ & \cellcolor{gray!10} $\mathbf{98.1}$ & \cellcolor{gray!10} $\bm{99.7}$ \\
& \cellcolor{gray!0} MAE$_{\text{rd}}$   & \cellcolor{gray!0} $77.3$ & \cellcolor{gray!0} $39.7$ & \cellcolor{gray!0} $79.6$ & \cellcolor{gray!0} $97.4$ & \cellcolor{gray!0} $99.6$ \\
& \cellcolor{gray!0} PMAE$_{\text{rd}}^{\star}$ & \cellcolor{gray!0} $\mathbf{80.4}$ & \cellcolor{gray!0} $\mathbf{46.9}$ & \cellcolor{gray!0} $\mathbf{82.4}$ & \cellcolor{gray!0} $\mathbf{98.5}$ & \cellcolor{gray!0} $\mathbf{99.7}$ \\
\bottomrule
\end{tabular}}
\vspace{-0.4cm}
\end{table}

\section{Results}\label{sec:results}
\looseness-1 In this section, we will outline and analyze the empirical advantages of PMAE compared to standard MAEs in image classification tasks. Specifically, we provide evidence that masking within the space of principal components facilitates the learning of discriminative features, resulting in improved classification performance. Our findings are supported by empirical evidence across five datasets, including two natural image datasets \rebuttal{of $32\!\times\!32$ and $64\!\times\!64$ resolutions}, and three medical datasets taken from MedMNIST \citep{yang2023medmnist} of $64\!\times\!64$ resolution (see \Cref{fig:medmnist} for examples of MedMNIST images).

\looseness-1 \Cref{tab:results} presents the classification accuracy using both a linear probe and a MLP classifier. Across datasets, we observe substantial improvements with PMAE$_{\text{ocl}}$ in linear probing compared to the standard MAE$_{\text{std}}$, with an average increase of 38\% ($+14$ percentage points). Additionally, PMAE$_{\text{ocl}}$ outperforms MAE$_{\text{ocl}}$ by 20.3\% ($+9.5$ percentage points). We observe similar trends with the randomized hyperparameter strategy: PMAE consistently outperforms MAE across all datasets, yielding an average performance increase of 29.9\% ($+5.4$ percentage points), even when sub-optimal hyperparameters are used. These findings also extend to the non-linear evaluation setting, (see the middle part of \Cref{tab:results}). The bottom part of \Cref{tab:results} presents the downstream performance in the fine-tuning setting. We continue to observe a performance gain brought by PMAE over MAE of 2.3\% ($+8.8$ percentage points) on average. For two of the MedMNIST datasets we observe an equal and high ($>\!98\%$) performance for both methods, showcasing that both approaches easily complete these tasks.

\looseness-1 These empirical findings lead to several conclusions. We observe that the recommended masking of $75\%$ of image patches is largely sub-optimal \emph{across} datasets. \Cref{fig:mae_hyperparameters} reports an ablation study of the masking ratios for MAE and PMAE. \Cref{fig:mae_hyperparameters} (top) shows that, across all five datasets, a $75\%$ masking ratio is sub-optimal. For PMAE, the masking ratio is a more stable hyperparameter. \Cref{fig:mae_hyperparameters} (bottom) shows that across all evaluated datasets we observe the best or near-optimal performance for PMAE at $20\%$ of the variance masked. We also validate the empirical benefits brought by PMAE. Interestingly, we notice that PMAE without any hyperparameter tuning (PMAE$_{\text{rd}}$) outperforms MAE with optimum masking ratio in all but one case (i.e., CIFAR10). Finally, masking 20\% of the variance in PMAE leads to substantial performance gains over the standard approach of using MAEs with a 75\% masking ratio. \

\begin{figure}  %
    \centering
    \vspace{-10pt}
    \includegraphics[width=0.4\textwidth]{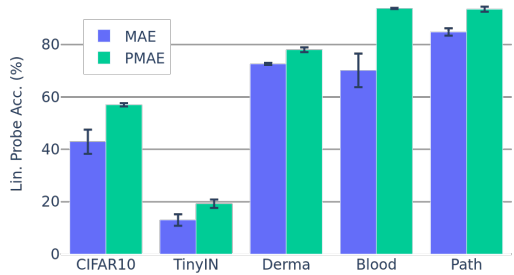}
    \vspace{-8pt}
    \caption{\small \textbf{Performance across masking ratios.} Average and standard error of the linear probe accuracy across masking ratios.}
    \vspace{-20pt}
    \label{fig:stability}
\end{figure}

\looseness-1 \Cref{fig:mae_hyperparameters} (right) shows the downstream performance over training epochs, where PMAE surpasses MAE as early as 200 epochs. Figures for other datasets are provided in \Cref{app:results}. Moreover, \Cref{fig:stability} examines how downstream accuracy varies with different masking ratios. Our findings indicate that PMAE demonstrates similar or lower standard errors across masking ratios compared to MAE. Overall, these results highlight that the masking strategy of PMAE better aligns image reconstruction with image classification tasks compared to the MAE objective.

\section{Understanding PMAE}\label{sec:explanation}
\looseness-1 In this section, we aim to provide more intuition as to why masking \textit{components} rather than \textit{image patches} leads to a more robust objective. In \cref{sec:mae}, we discuss the hypothesis under which MIM operates \citep{kong2023understanding} and present an example failure case of spatial masking in \Cref{fig:pixel_mask}. We highlight how masking pixels can lead to a misalignment between the MIM objective and the learning of meaningful representations. If all patches covering an object are masked out, it is uncertain whether the remaining patches share any information with the object. On the contrary, if the masked-out information is redundant with the information carried by visible patches, it is likely that the information shared does not contain the object class but rather perceptual features (e.g., colors or textures).

\begin{figure*}[t]
    \centering
    \includegraphics[width=\linewidth]{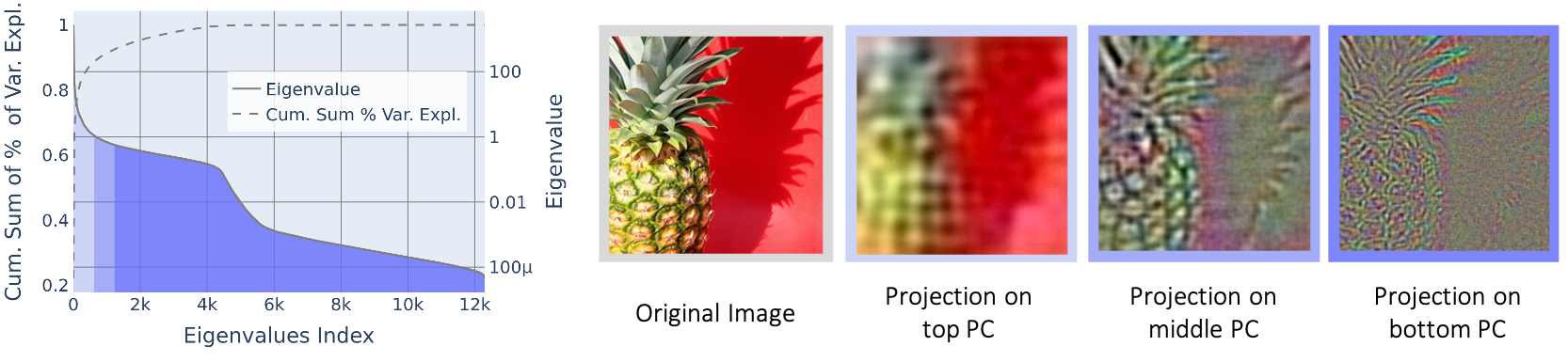} 
    \vspace{-0.8cm}
    \caption{\small \textbf{From Principal Components to Spatial Features.} Overview of the spatial features associated with distinct regions of the principal component spectrum; Images depict the features captured by the top (light blue), middle (mild blue) and bottom (dark blue) PCs.}
    \vspace{-0.3cm}
    \label{fig:masking_pc}
\end{figure*}

\looseness-1 Different from spatial masking, masking principal \textit{components} leads to the removal of global image features, instead of only acting locally as in spatial masking. \Cref{fig:masking_pc} serves as an example highlighting the correspondence between principal component and perceptual features. In this example, the principal components with the highest eigenvalues capture the colors within the image while the bottom PCs highlight the edges. Early work in image processing \citep{turk1991} has demonstrated this connection between an image's dominant modes of variation and its low spatial frequency components, providing further intuition for how information is partitioned in the space of PCs of natural images.

\looseness-1 By removing a subset of principal components, PMAE prevents the removal of all the information characterizing an object and prevents redundant information to remain after masking. Instead, PMAE drops a set of unique image components. By taking advantage of the information partitioning in PCA, PMAE thereby mitigates MAE's failure cases, ultimately leading to increased accuracy. Although the potential of the principal component space for MIM \citep{balestriero_learning_2024} or Image Denoising \citep{chen2024deconstructing} has been recently explored, our work is the first to propose an effective masking strategy that directly leverages PCA.

\section{Related Work}
\textbf{Self-supervised learning.} 
\looseness-1 
Self-supervised learning (SSL) leverages auxiliary tasks to learn from unlabeled data, often outperforming supervised methods on downstream tasks. SSL can be divided into two categories: discriminative and generative \citep{liu2021self}. Discriminative methods \citep{chen2020simple,caron_emerging_2021} focus on enforcing invariance or equivariance between data views in the representation space, while generative methods \citep{he_masked_2021,bizeul2024probabilistic} rely on data reconstruction from, often, corrupted observations. Though generative SSL historically lagged in performance, recent work has bridged the gap by integrating strengths from both paradigms \citep{assran_masked_2022,dong2023peco,oquab2023dinov2,chen2024context,lehner_contrastive_2023}. Interestingly, recent discriminative methods employ cropping strategies to create distinct data views \citep{oquab2023dinov2,assran_self-supervised_2023}, which is reminiscent of image masking. \citet{balestriero_learning_2024} point out the misalignment between auxiliary and downstream tasks in reconstruction-based SSL and suggest novel masking strategies to help realign these objectives.

\textbf{Masked Image Modelling.} \looseness-1 MIM extends the successful masked language modeling paradigm to vision tasks. Early methods, such as Context Encoder \citep{pathak2016context}, used a convolutional autoencoder to inpaint a central region of the image. The rise of Vision Transformers (ViTs) \citep{dosovitskiy_image_2021} has driven significant advancements in MIM. BEiT \citep{zhou2021ibot,bao_beit_2022} combines a ViT encoder with image tokenizers \citep{ramesh2021zero} to predict discrete tokens for masked patches. SimMIM \citep{xie2022simmim} simplifies the task by pairing a ViT encoder with a regression head to directly predict raw pixel values for the masked regions. MAE \citep{he_masked_2021} introduces a more efficient encoder-decoder architecture, with a shallow decoder. MIM’s domain-agnostic masking strategies have also proven effective in multi-modal tasks \citep{baevski_data2vec_2022,bachmann2022multimae}.

\textbf{Mask Design Strategies.} \looseness-1 A critical component of the masked image modeling paradigm is the design of effective masking strategies. Early MIM approaches have relied on random spatial masking techniques, such as masking out the central region of an image \citep{pathak2016context}, image patches \citep{he_masked_2021,xie2022simmim}, and blocks of patches \citep{bao_beit_2022}. Inspired by advances in language modeling, recent efforts have explored semantically guided mask design. \citet{li_mst_2021} use self-attention maps to mask irrelevant regions, while \citet{kakogeorgiou_what_2022} focus on masking semantically rich areas. \citet{shi_adversarial_2022} design masks through adversarial learning, where the resulting masks resemble semantic maps, a concept extended by \citet{li_semmae_2022} through progressive semantic region masking. Further advancing this direction, \citet{wang_hard_2023} and \citet{madan_cl-mae_2024} introduce curriculum learning-inspired mask design methods. %
These methods often require additional training steps, components, or more complex objectives. More closely related to our work, \citet{chang2022maskgit, chen2024deconstructing} explore the use of pre-existing image representations for asked Image Modeling and image denoising. \citet{chen2024deconstructing} introduce additive Gaussian noise to principal components as an alternative to the traditional Denoising Autoencoders. \citet{chang2022maskgit} utilize masked token modeling by leveraging the discrete latent space of a pre-trained VQVAE to develop an image generation model.

\section{Discussion}\label{sec:discussion}
\looseness-1 In this work, we have investigated different masking strategies for Masked Image Modelling (MIM). To this end, we have introduced the Principal Masked Autoencoder (PMAE) as an alternative to spatial masking. PMAE is rooted in principal component analysis ~\citep[PCA;][]{pearson1901,hotelling1933analysis}, which is a widely used data-driven, \textit{linear} and deterministic transformation.
Unlike recent alternatives that require additional supervision, learnable components, or complex training pipelines \citep{li_mst_2021,li_semmae_2022,kakogeorgiou_what_2022,li_progressively_2022}, PMAE stays close to the core principles of MIM: the combination of a randomized masking strategy and an encoder-decoder architecture. Despite its simplicity, we demonstrate that PMAE yields substantial performance improvements over spatial masking on image classification tasks for CIFAR10, TinyImageNet and MedMNIST datasets. Further, in a PMAE, the masking ratio---typically a sensitive and difficult-to-tune hyperparameter in MIM---appears more robust and has a natural interpretation as the ratio of variance explained by the masked input. 

\looseness-1 Since PCA, or its generalisation kernel PCA~\citep{scholkopf1997kernel}, is easily applicable to any data modality,
our proposal of masking principal components is not specific to MIM. Instead, it can be viewed as a \textit{general strategy} that should also be applicable to other types of modalities beyond images, as well as to other self-supervised learning (SSL) approaches. Indeed, data masking is commonly adopted in discriminative SSL methods. Whereas early approaches, such as SimCLR~\citep{chen2020simple} or MoCo~\citep{chen2020improved}, relied on combinations of image transformations (e.g., color jitter, flips, crops, etc.) as data augmentation strategies, more recent state-of-the-art methods like DINO~\citep{caron_emerging_2021,oquab2023dinov2} and I-JEPA~\citep{assran_self-supervised_2023} have shifted to relying solely on image cropping, which can be considered a type of masking. The integration of principal component masking instead of image cropping into such SSL pipelines constitutes a promising future direction of research. 

\looseness-1  In the present work, we have focused on PCA subspaces as meaningful masking spaces. However, our core
idea of masking a \textit{transformed} version of the data (rather than the \textit{raw} data) can be viewed as laying
the groundwork for other, more generic approaches to information masking for self-supervised representation learning. Moving beyond PCA, a natural extension would be to \textit{learn} a suitable latent space in which the masking is performed.
This route can potentially leverage recent theoretical insights~\citep{kong2023understanding} by more explicitly enforcing that the shared information between visible and masked-out latent components contains high-level latent variables that are most useful for the downstream tasks of interest. Other off-the-shelf transformations, such as the Fourier transform \citep{bracewell1966fourier}, Wavelet transform \citep{daubechies1992ten}, Kernel principal component analysis \citep{scholkopf1997kernel}, or Diffusion Maps \citep{coifman2006diffusion}, represent alternative candidate transformations. Future research should explore whether the properties of these spaces provide comparable or additional advantages over PCA. Preliminary results with non-linear transformations, namely Kernel PCA, presented in \Cref{app:kernel}, demonstrate performance gains over PMAE, motivating further exploration. A particularly appealing aspect of some of these methods (e.g., Fourier \& Wavelet transforms and Diffusion Maps) is the use of fixed bases, which could eliminate the computational overhead of PCA---whose cost scales cubically with the data dimensionality---and improve scalability to larger datasets.

\looseness-1 Our work demonstrates the potential of using PCA subspaces to guide image representation learning. Exploring alternative masking spaces with similar properties presents an exciting avenue for future research, offering the possibility of developing unsupervised tasks that encourage the learning of features more closely aligned with human perception.
 
\section*{Acknowledgments} 
\looseness-1 We thank Randall Balestreiro, Carl Allen, Maximilian Dax, Georgios Kissas, and David Klindt for useful discussions and comments. Alice Bizeul is supported by an ETH AI Center Doctoral Fellowship. Alain Ryser is supported by the StimuLoop grant \#1-007811-002 and the Vontobel Foundation. 

\bibliography{pcmae}

\begin{thebibliography}{45}
\providecommand{\natexlab}[1]{#1}
\providecommand{\url}[1]{\texttt{#1}}
\expandafter\ifx\csname urlstyle\endcsname\relax
  \providecommand{\doi}[1]{doi: #1}\else
  \providecommand{\doi}{doi: \begingroup \urlstyle{rm}\Url}\fi

\bibitem[Assran et~al.(2022)Assran, Caron, Misra, Bojanowski, Bordes, Vincent, Joulin, Rabbat, and Ballas]{assran_masked_2022}
Assran, M., Caron, M., Misra, I., Bojanowski, P., Bordes, F., Vincent, P., Joulin, A., Rabbat, M., and Ballas, N.
\newblock Masked siamese networks for label-efficient learning.
\newblock In \emph{European Conference on Computer Vision}, 2022.
\newblock URL \url{http://arxiv.org/abs/2204.07141}.

\bibitem[Assran et~al.(2023)Assran, Duval, Misra, Bojanowski, Vincent, Rabbat, {LeCun}, and Ballas]{assran_self-supervised_2023}
Assran, M., Duval, Q., Misra, I., Bojanowski, P., Vincent, P., Rabbat, M., {LeCun}, Y., and Ballas, N.
\newblock Self-supervised learning from images with a joint-embedding predictive architecture.
\newblock In \emph{Proceedings of the {IEEE}/{CVF} Conference on Computer Vision and Pattern Recognition}, 2023.
\newblock URL \url{https://arxiv.org/abs/2301.08243}.

\bibitem[Bachmann et~al.(2022)Bachmann, Mizrahi, Atanov, and Zamir]{bachmann2022multimae}
Bachmann, R., Mizrahi, D., Atanov, A., and Zamir, A.
\newblock Multimae: Multi-modal multi-task masked autoencoders.
\newblock In \emph{European Conference on Computer Vision}, 2022.
\newblock URL \url{https://arxiv.org/pdf/2204.01678}.

\bibitem[Baevski et~al.(2022)Baevski, Hsu, Xu, Babu, Gu, and Auli]{baevski_data2vec_2022}
Baevski, A., Hsu, W.-N., Xu, Q., Babu, A., Gu, J., and Auli, M.
\newblock data2vec: A general framework for self-supervised learning in speech, vision and language.
\newblock In \emph{Proceedings of the 39th International Conference on Machine Learning}, 2022.
\newblock URL \url{https://arxiv.org/abs/2202.03555}.

\bibitem[Balestriero \& LeCun(2024)Balestriero and LeCun]{balestriero_learning_2024}
Balestriero, R. and LeCun, Y.
\newblock How learning by reconstruction produces uninformative features for perception.
\newblock In \emph{Forty-first International Conference on Machine Learning}, 2024.
\newblock URL \url{http://arxiv.org/abs/2402.11337}.

\bibitem[Bao et~al.(2022)Bao, Dong, Piao, and Wei]{bao_beit_2022}
Bao, H., Dong, L., Piao, S., and Wei, F.
\newblock {BEiT}: {BERT} pre-training of image transformers.
\newblock In \emph{International Conference on Learning Representations}, 2022.
\newblock URL \url{http://arxiv.org/abs/2106.08254}.

\bibitem[Bengio et~al.(2013)Bengio, Courville, and Vincent]{bengio2013representation}
Bengio, Y., Courville, A., and Vincent, P.
\newblock Representation learning: A review and new perspectives.
\newblock \emph{IEEE transactions on pattern analysis and machine intelligence}, 2013.
\newblock URL \url{https://arxiv.org/pdf/1206.5538}.

\bibitem[Bizeul et~al.(2024)Bizeul, Sch{\"o}lkopf, and Allen]{bizeul2024probabilistic}
Bizeul, A., Sch{\"o}lkopf, B., and Allen, C.
\newblock A probabilistic model to explain self-supervised representation learning.
\newblock \emph{Transactions on Machine Learning Research}, 2024.
\newblock URL \url{https://arxiv.org/pdf/2402.01399}.

\bibitem[Bracewell \& Kahn(1966)Bracewell and Kahn]{bracewell1966fourier}
Bracewell, R. and Kahn, P.~B.
\newblock The fourier transform and its applications.
\newblock \emph{American Journal of Physics}, 1966.

\bibitem[Caron et~al.(2021)Caron, Touvron, Misra, Jégou, Mairal, Bojanowski, and Joulin]{caron_emerging_2021}
Caron, M., Touvron, H., Misra, I., Jégou, H., Mairal, J., Bojanowski, P., and Joulin, A.
\newblock Emerging properties in self-supervised vision transformers.
\newblock In \emph{Proceedings of the IEEE/CVF international conference on computer vision}, 2021.
\newblock URL \url{http://arxiv.org/abs/2104.14294}.

\bibitem[Chang et~al.(2022)Chang, Zhang, Jiang, Liu, and Freeman]{chang2022maskgit}
Chang, H., Zhang, H., Jiang, L., Liu, C., and Freeman, W.~T.
\newblock Maskgit: Masked generative image transformer.
\newblock In \emph{Proceedings of the IEEE/CVF Conference on Computer Vision and Pattern Recognition}, 2022.
\newblock URL \url{https://arxiv.org/abs/2202.04200}.

\bibitem[Chen et~al.(2020{\natexlab{a}})Chen, Kornblith, Norouzi, and Hinton]{chen2020simple}
Chen, T., Kornblith, S., Norouzi, M., and Hinton, G.
\newblock A simple framework for contrastive learning of visual representations.
\newblock In \emph{International conference on machine learning}, 2020{\natexlab{a}}.
\newblock URL \url{http://proceedings.mlr.press/v119/chen20j/chen20j.pdf}.

\bibitem[Chen et~al.(2020{\natexlab{b}})Chen, Fan, Girshick, and He]{chen2020improved}
Chen, X., Fan, H., Girshick, R., and He, K.
\newblock Improved baselines with momentum contrastive learning.
\newblock \emph{arXiv preprint arXiv:2003.04297}, 2020{\natexlab{b}}.
\newblock URL \url{https://arxiv.org/abs/2003.04297}.

\bibitem[Chen et~al.(2024{\natexlab{a}})Chen, Ding, Wang, Xin, Mo, Wang, Han, Luo, Zeng, and Wang]{chen2024context}
Chen, X., Ding, M., Wang, X., Xin, Y., Mo, S., Wang, Y., Han, S., Luo, P., Zeng, G., and Wang, J.
\newblock Context autoencoder for self-supervised representation learning.
\newblock \emph{International Journal of Computer Vision}, 132\penalty0 (1):\penalty0 208--223, 2024{\natexlab{a}}.
\newblock URL \url{https://link.springer.com/content/pdf/10.1007/s11263-023-01852-4.pdf}.

\bibitem[Chen et~al.(2024{\natexlab{b}})Chen, Liu, Xie, and He]{chen2024deconstructing}
Chen, X., Liu, Z., Xie, S., and He, K.
\newblock Deconstructing denoising diffusion models for self-supervised learning.
\newblock \emph{arXiv preprint arXiv:2401.14404}, 2024{\natexlab{b}}.
\newblock URL \url{https://arxiv.org/pdf/2401.14404}.

\bibitem[Coifman \& Lafon(2006)Coifman and Lafon]{coifman2006diffusion}
Coifman, R.~R. and Lafon, S.
\newblock Diffusion maps.
\newblock \emph{Applied and computational harmonic analysis}, 2006.
\newblock URL \url{https://www.sciencedirect.com/science/article/pii/S1063520306000546}.

\bibitem[Daubechies(1992)]{daubechies1992ten}
Daubechies, I.
\newblock Ten lectures on wavelets.
\newblock \emph{Society for industrial and applied mathematics}, 1992.
\newblock URL \url{https://epubs.siam.org/doi/pdf/10.1137/1.9781611970104.fm}.

\bibitem[Devlin(2019)]{devlin2018bert}
Devlin, J.
\newblock Bert: Pre-training of deep bidirectional transformers for language understanding.
\newblock In \emph{Proceedings of the 2019 Conference of the North American Chapter of the Association for Computational Linguistics: Human Language Tech- nologies}, 2019.
\newblock URL \url{https://arxiv.org/abs/1810.04805}.

\bibitem[Dong et~al.(2023)Dong, Bao, Zhang, Chen, Zhang, Yuan, Chen, Wen, Yu, and Guo]{dong2023peco}
Dong, X., Bao, J., Zhang, T., Chen, D., Zhang, W., Yuan, L., Chen, D., Wen, F., Yu, N., and Guo, B.
\newblock Peco: Perceptual codebook for bert pre-training of vision transformers.
\newblock In \emph{Proceedings of the AAAI Conference on Artificial Intelligence}, 2023.
\newblock URL \url{https://arxiv.org/pdf/2111.12710}.

\bibitem[Dosovitskiy et~al.(2021)Dosovitskiy, Beyer, Kolesnikov, Weissenborn, Zhai, Unterthiner, Dehghani, Minderer, Heigold, Gelly, Uszkoreit, and Houlsby]{dosovitskiy_image_2021}
Dosovitskiy, A., Beyer, L., Kolesnikov, A., Weissenborn, D., Zhai, X., Unterthiner, T., Dehghani, M., Minderer, M., Heigold, G., Gelly, S., Uszkoreit, J., and Houlsby, N.
\newblock An image is worth 16x16 words: Transformers for image recognition at scale.
\newblock In \emph{Proceedings of the Tenth International Conference on Learning Representations.}, 2021.
\newblock URL \url{http://arxiv.org/abs/2010.11929}.

\bibitem[Goyal et~al.(2017)Goyal, Doll{\'{a}}r, Girshick, Noordhuis, Wesolowski, Kyrola, Tulloch, Jia, and He]{goyal2017accurate}
Goyal, P., Doll{\'{a}}r, P., Girshick, R.~B., Noordhuis, P., Wesolowski, L., Kyrola, A., Tulloch, A., Jia, Y., and He, K.
\newblock Accurate, large minibatch sg d: training imagenet in 1 hour.
\newblock \emph{arXiv preprint arXiv:1706.02677}, 2017.
\newblock URL \url{https://arxiv.org/abs/1706.02677}.

\bibitem[He et~al.(2021)He, Chen, Xie, Li, Dollár, and Girshick]{he_masked_2021}
He, K., Chen, X., Xie, S., Li, Y., Dollár, P., and Girshick, R.
\newblock Masked autoencoders are scalable vision learners.
\newblock In \emph{Proceedings of the IEEE/CVF conference on computer vision and pattern recognition}, 2021.
\newblock URL \url{http://arxiv.org/abs/2111.06377}.

\bibitem[Hotelling(1933)]{hotelling1933analysis}
Hotelling, H.
\newblock Analysis of a complex of statistical variables into principal components.
\newblock \emph{Journal of educational psychology}, 1933.
\newblock URL \url{https://www.cis.rit.edu/~rlepci/Erho/Derek/Useful_References/Principal\%20Components\%20Analysis/Hotelling_PCA_part1.pdf}.

\bibitem[Kakogeorgiou et~al.(2022)Kakogeorgiou, Gidaris, Psomas, Avrithis, Bursuc, Karantzalos, and Komodakis]{kakogeorgiou_what_2022}
Kakogeorgiou, I., Gidaris, S., Psomas, B., Avrithis, Y., Bursuc, A., Karantzalos, K., and Komodakis, N.
\newblock What to hide from your students: Attention-guided masked image modeling.
\newblock In \emph{ECCV}, 2022.
\newblock URL \url{https://link.springer.com/chapter/10.1007/978-3-031-20056-4_18}.

\bibitem[Kong et~al.(2023)Kong, Ma, Chen, Xing, Chi, Morency, and Zhang]{kong2023understanding}
Kong, L., Ma, M.~Q., Chen, G., Xing, E.~P., Chi, Y., Morency, L.-P., and Zhang, K.
\newblock Understanding masked autoencoders via hierarchical latent variable models.
\newblock In \emph{Proceedings of the IEEE/CVF Conference on Computer Vision and Pattern Recognition}, 2023.
\newblock URL \url{https://arxiv.org/abs/2306.04898}.

\bibitem[Lehner et~al.(2023)Lehner, Alkin, Fürst, Rumetshofer, Miklautz, and Hochreiter]{lehner_contrastive_2023}
Lehner, J., Alkin, B., Fürst, A., Rumetshofer, E., Miklautz, L., and Hochreiter, S.
\newblock Contrastive tuning: A little help to make masked autoencoders forget.
\newblock In \emph{Proceedings of the AAAI Conference on Artificial Intelligence}, 2023.
\newblock URL \url{http://arxiv.org/abs/2304.10520}.

\bibitem[Li et~al.(2022{\natexlab{a}})Li, Zheng, Liu, Wang, Su, and Zheng]{li_semmae_2022}
Li, G., Zheng, H., Liu, D., Wang, C., Su, B., and Zheng, C.
\newblock {SemMAE}: Semantic-guided masking for learning masked autoencoders.
\newblock \emph{Advances in Neural Information Processing Systems}, 2022{\natexlab{a}}.
\newblock URL \url{https://arxiv.org/abs/2206.10207}.

\bibitem[Li et~al.(2022{\natexlab{b}})Li, Wang, Zhang, Chen, Jiang, Dai, Li, Xiong, and Tian]{li_progressively_2022}
Li, J., Wang, Y., Zhang, X., Chen, Y., Jiang, D., Dai, W., Li, C., Xiong, H., and Tian, Q.
\newblock Progressively compressed auto-encoder for self-supervised representation learning.
\newblock In \emph{The Eleventh International Conference on Learning Representations}, 2022{\natexlab{b}}.
\newblock URL \url{https://openreview.net/pdf?id=8T4qmZbTkW7}.

\bibitem[Li et~al.(2021)Li, Chen, Yang, Li, Zhu, Zhao, Deng, Wu, Zhao, Tang, and Wang]{li_mst_2021}
Li, Z., Chen, Z., Yang, F., Li, W., Zhu, Y., Zhao, C., Deng, R., Wu, L., Zhao, R., Tang, M., and Wang, J.
\newblock {MST}: Masked self-supervised transformer for visual representation.
\newblock \emph{Advances in Neural Information Processing Systems}, 2021.
\newblock URL \url{http://arxiv.org/abs/2106.05656}.

\bibitem[Liu et~al.(2021)Liu, Zhang, Hou, Mian, Wang, Zhang, and Tang]{liu2021self}
Liu, X., Zhang, F., Hou, Z., Mian, L., Wang, Z., Zhang, J., and Tang, J.
\newblock Self-supervised learning: Generative or contrastive.
\newblock \emph{IEEE transactions on knowledge and data engineering}, 2021.
\newblock URL \url{https://arxiv.org/pdf/2006.08218}.

\bibitem[Madan et~al.(2024)Madan, Ristea, Nasrollahi, Moeslund, and Ionescu]{madan_cl-mae_2024}
Madan, N., Ristea, N.-C., Nasrollahi, K., Moeslund, T.~B., and Ionescu, R.~T.
\newblock {CL}-{MAE}: Curriculum-learned masked autoencoders.
\newblock In \emph{Proceedings of the {IEEE}/{CVF} Winter Conference on Applications of Computer Vision}, 2024.
\newblock URL \url{https://arxiv.org/abs/2308.16572}.

\bibitem[Oquab et~al.(2023)Oquab, Darcet, Moutakanni, Vo, Szafraniec, Khalidov, Fernandez, Haziza, Massa, El-Nouby, et~al.]{oquab2023dinov2}
Oquab, M., Darcet, T., Moutakanni, T., Vo, H., Szafraniec, M., Khalidov, V., Fernandez, P., Haziza, D., Massa, F., El-Nouby, A., et~al.
\newblock Dinov2: Learning robust visual features without supervision.
\newblock \emph{TMLR}, 2023.
\newblock URL \url{https://arxiv.org/pdf/2304.07193}.

\bibitem[Pathak et~al.(2016)Pathak, Krahenbuhl, Donahue, Darrell, and Efros]{pathak2016context}
Pathak, D., Krahenbuhl, P., Donahue, J., Darrell, T., and Efros, A.~A.
\newblock Context encoders: Feature learning by inpainting.
\newblock In \emph{Proceedings of the IEEE conference on computer vision and pattern recognition}, 2016.
\newblock URL \url{https://arxiv.org/pdf/1604.07379}.

\bibitem[Pearson(1901)]{pearson1901}
Pearson, K.
\newblock Liii. on lines and planes of closest fit to systems of points in space.
\newblock \emph{The London, Edinburgh, and Dublin Philosophical Magazine and Journal of Science}, 1901.
\newblock URL \url{https://doi.org/10.1080/14786440109462720}.

\bibitem[Ramesh et~al.(2021)Ramesh, Pavlov, Goh, Gray, Voss, Radford, Chen, and Sutskever]{ramesh2021zero}
Ramesh, A., Pavlov, M., Goh, G., Gray, S., Voss, C., Radford, A., Chen, M., and Sutskever, I.
\newblock Zero-shot text-to-image generation.
\newblock In \emph{International conference on machine learning}, 2021.
\newblock URL \url{http://proceedings.mlr.press/v139/ramesh21a/ramesh21a.pdf}.

\bibitem[Sch{\"o}lkopf et~al.(1997)Sch{\"o}lkopf, Smola, and M{\"u}ller]{scholkopf1997kernel}
Sch{\"o}lkopf, B., Smola, A., and M{\"u}ller, K.-R.
\newblock Kernel principal component analysis.
\newblock In \emph{International conference on artificial neural networks}, 1997.
\newblock URL \url{https://link.springer.com/chapter/10.1007/BFb0020217}.

\bibitem[Shi et~al.(2022)Shi, Siddharth, Torr, and Kosiorek]{shi_adversarial_2022}
Shi, Y., Siddharth, N., Torr, P., and Kosiorek, A.~R.
\newblock Adversarial masking for self-supervised learning.
\newblock In \emph{Proceedings of the 39th International Conference on Machine Learning}, 2022.
\newblock URL \url{https://arxiv.org/abs/2201.13100}.

\bibitem[Touvron et~al.(2021)Touvron, Cord, Douze, Massa, Sablayrolles, and J{\'e}gou]{touvron2021training}
Touvron, H., Cord, M., Douze, M., Massa, F., Sablayrolles, A., and J{\'e}gou, H.
\newblock Training data-efficient image transformers \& distillation through attention.
\newblock In \emph{International conference on machine learning}, 2021.
\newblock URL \url{https://arxiv.org/abs/2012.12877}.

\bibitem[Turk \& Pentland(1991)Turk and Pentland]{turk1991}
Turk, M. and Pentland, A.
\newblock {Eigenfaces for Recognition}.
\newblock \emph{Journal of Cognitive Neuroscience}, 1991.
\newblock URL \url{https://doi.org/10.1162/jocn.1991.3.1.71}.

\bibitem[Wang et~al.(2023)Wang, Song, Fan, Wang, Xie, and Zhang]{wang_hard_2023}
Wang, H., Song, K., Fan, J., Wang, Y., Xie, J., and Zhang, Z.
\newblock Hard patches mining for masked image modeling.
\newblock In \emph{Proceedings of the {IEEE}/{CVF} Conference on Computer Vision and Pattern Recognition}, 2023.
\newblock URL \url{https://arxiv.org/abs/2304.05919}.

\bibitem[Xie et~al.(2022)Xie, Zhang, Cao, Lin, Bao, Yao, Dai, and Hu]{xie2022simmim}
Xie, Z., Zhang, Z., Cao, Y., Lin, Y., Bao, J., Yao, Z., Dai, Q., and Hu, H.
\newblock Simmim: A simple framework for masked image modeling.
\newblock In \emph{Proceedings of the IEEE/CVF conference on computer vision and pattern recognition}, 2022.
\newblock URL \url{https://arxiv.org/abs/2111.09886}.

\bibitem[Yang et~al.(2023)Yang, Shi, Wei, Liu, Zhao, Ke, Pfister, and Ni]{yang2023medmnist}
Yang, J., Shi, R., Wei, D., Liu, Z., Zhao, L., Ke, B., Pfister, H., and Ni, B.
\newblock Medmnist v2-a large-scale lightweight benchmark for 2d and 3d biomedical image classification.
\newblock \emph{Scientific Data}, 2023.
\newblock URL \url{https://www.nature.com/articles/s41597-022-01721-8}.

\bibitem[Yue et~al.(2023)Yue, Bai, Wei, Pang, Liu, Zhou, and Ouyang]{yue_understanding_2023}
Yue, X., Bai, L., Wei, M., Pang, J., Liu, X., Zhou, L., and Ouyang, W.
\newblock Understanding masked autoencoders from a local contrastive perspective.
\newblock \emph{arXiv preprint arXiv:2310.01994}, 2023.
\newblock URL \url{http://arxiv.org/abs/2310.01994}.

\bibitem[Zhang et~al.(2022)Zhang, Wang, and Wang]{zhang_how_2022}
Zhang, Q., Wang, Y., and Wang, Y.
\newblock How mask matters: {Towards} theoretical understandings of masked autoencoders.
\newblock \emph{Advances in Neural Information Processing Systems}, 2022.
\newblock URL \url{https://arxiv.org/abs/2210.08344}.

\bibitem[Zhou et~al.(2021)Zhou, Wei, Wang, Shen, Xie, Yuille, and Kong]{zhou2021ibot}
Zhou, J., Wei, C., Wang, H., Shen, W., Xie, C., Yuille, A., and Kong, T.
\newblock ibot: Image bert pre-training with online tokenizer.
\newblock In \emph{International Conference on Learning Representations}, 2021.
\newblock URL \url{https://arxiv.org/pdf/2111.07832}.

\end{thebibliography}
\bibliographystyle{icml2025}

\newpage
\appendix
\onecolumn
\section{Appendix}

\subsection{Experimental Setup}\label{app:setup}
\subsubsection{Datasets}
\textbf{CIFAR-10} is a widely used benchmark dataset containing 50,000 training and 10,000 validation $32\!\times\!32$ RGB images depicting 10 object classes, such as airplanes, cars, and animals.

\begin{wrapfigure}{r}{0.5\textwidth}
    \centering
    \vspace{-5pt}
    \includegraphics[width=0.48\textwidth]{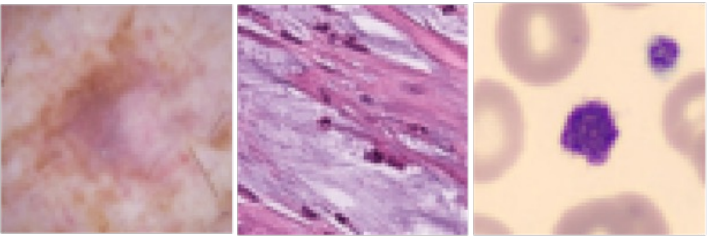}
    \caption{\small{ \textbf{MedMNIST datasets.} Example images from the (from left to right) DermaMNIST, 
    PathMNIST, and BloodMNIST datasets used for image classification.}}
    \label{fig:medmnist}
    \vspace{-10pt}
\end{wrapfigure}

\textbf{TinyImageNet} is a subset of the ImageNet dataset, containing 200 classes of $64\!\times\!64$ RGB images. It consists of 100,000 training images and 10,000 validation images, making it a challenging benchmark for classification tasks with more fine-grained object categories compared to CIFAR-10. Examples images are depicted in \Cref{fig:medmnist}

The MedMNIST \citep{yang2023medmnist} dataset is a collection of medical imaging datasets, each focusing on different types of biomedical data. In this work, three MedMNIST datasets are used:

\textbf{BloodMNIST} \looseness-1 consists of 12,000 training and 1,700 validation $64\!\times\!64$ RGB images across 8 classes and represents microscopic images of blood cells, used for hematology classification tasks.

\textbf{DermaMNIST} \looseness-1 contains 7,000 training and 1,000 validation $64\!\times\!64$ RGB images of skin samples, each part of one out of 7 types of skin diseases.

\textbf{PathMNIST} \looseness-1 comprises 90,000 training and 10,000 validation $64\!\times\!64$ RGB images across 9 classes and depicts histopathological images of colorectal cancer tissue, aiding in classification tasks relevant to pathology.

\looseness-1 We apply an equivalent data augmentation strategy to all datasets and for all learning objectives during training; Following \citet{he_masked_2021}, our augmentation strategy consists of a random cropping followed by image resizing using bicubic interpolation. The scale of the random cropping is fixed to $[0.2,1.0]$. We add horizontal flipping and we normalize images using each dataset's training mean and standard deviation; We keep these data augmentations during training and evaluation. We find that the use of data augmentations during the evaluation leads to a substantial performance drop for PMAE but we do keep these augmentations for fair comparisons. For all datasets and methods, we define image patches as patches of $8\!\times\!8$ pixels.

\begin{figure*}[ht]
    \centering
    \vspace{-0.1cm}
    \includegraphics[width=1.01\linewidth]{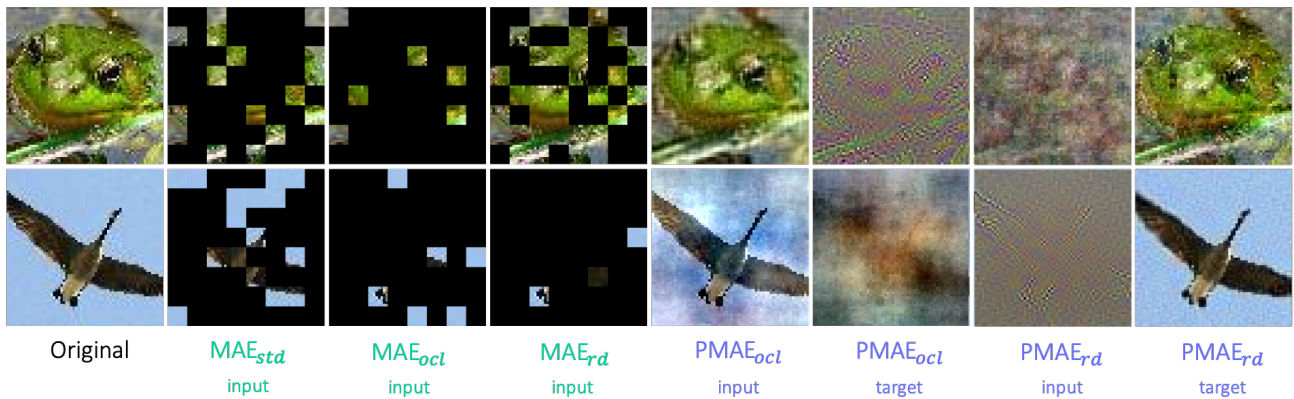}
    \vspace{-0.1cm}
    \caption{\small \textbf{Mask Design Strategies.} An overview of the different mask design strategies used in our experimental setup: spatial masking (\textit{green}) and principal component masking (\textit{blue}). \textit{std} refers to the standard approach of masking out $75\%$ of image patches, \textit{ocl} denotes masking with the optimal masking ratio, \textit{rd} represents a randomized strategy where the masking ratio is randomly sampled for each batch, and \textit{target} refers to the reconstruction target.}
    \label{fig:masking}
\end{figure*}

\subsubsection{Mask Design}
\Cref{fig:masking} depicts examples of the different masking strategies explored in \Cref{sec:results}. For MAE with \textit{std} masking, the masking ratio is fixed to 75--- 75\% of patches are masked out. For MAE and PMAE with \textit{ocl} masking, we tune the masking ratio to find the optimal ratio for each dataset and method. For \textit{rd} masking, the masking ratio is sampled uniformly from the range [10,90] for each new batch of data during training. 

\begin{wrapfigure}[7]{r}{0.33\textwidth}  %
    \vspace{-1.5cm}  %
    \centering
    \begin{minipage}{0.3\textwidth}  %
        \centering
        \begin{tabular}{ll}
        \toprule
        \textbf{config}              & \textbf{value} \\
        \midrule
        hidden size                  & 192            \\
        number of attention heads    & 12             \\
        intermediate size            & 768            \\
        norm pixel loss              & Y/N            \\
        patch size                   & $8\!\times\!8$ \\
        \bottomrule
        \end{tabular}
        \vspace{-5pt}
        \caption{ViT-T/8 hyperparameters.}
        \label{app_tab:model}
    \end{minipage}
    \vspace{-0.5cm}
\end{wrapfigure}

\subsubsection{Model Architecture}
We train a tiny Vision Transformer encoder architecture (ViT-T) with image patch size $8\!\times\!8$ for all datasets (ViT-T/8). The specifics of this architecture can be found in \Cref{app_tab:model}. Note that while for MAE, we keep the normalized pixel values of each masked patch as reconstruction targets \citep{he_masked_2021}, we find it to not have a clear positive impact for PMAE and hence enforce the reconstruction of the raw target values for PMAE.

\subsubsection{Training Hyperparameters}
We train the ViT-T encoder-decoder architecture for 800 epochs with the hyperparameters found in \Cref{app_tab:training}. These hyperparameters are taken from \citep{he_masked_2021}. We use the linear lr scaling rule: lr = base lr×batchsize / 256 \citep{goyal2017accurate}. Note that for our oracle masking settings, we conduct ablation studies across a masking ratio range of $[10,90]$.

\subsubsection{Evaluation Hyperparameters}
We evaluate the learned representation (i.e., \texttt{[CLS]} token) using a linear probe, multi-layer perceptron (MLP) classifier, and $k$-Nearest Neighbors algorithm on top of the frozen representation. The training samples of each dataset are used for training and the validation samples for testing. For linear and MLP probing experiments, we train the probes for 100 epochs following common practices \citep{he_masked_2021}. For the $k$-NN algorithm, we tune the number of neighbors in the range $[2,20]$. More details regarding the linear probing evaluation hyperparameters can be found in \Cref{app_tab:evaluation_lin}. The same hyperparameters were used for the MLP probing. We also use the linear lr scaling rule: lr = base lr×batchsize / 256 \citep{goyal2017accurate}. For the fine-tuning setup, we also follow \citep{he_masked_2021}, we fine-tune the encoder and a linear probe for 100 epochs and resort to using the hyperparameters presented in \Cref{app_tab:evaluation_fine}. Note that for PMAE, we evaluate the approach on raw images and do not perform any filtering of principal components prior to evaluation.

\subsubsection{Computational Resources}

\begin{figure}
        \centering
    \vspace{-5pt}
    \includegraphics[width=0.6\textwidth]{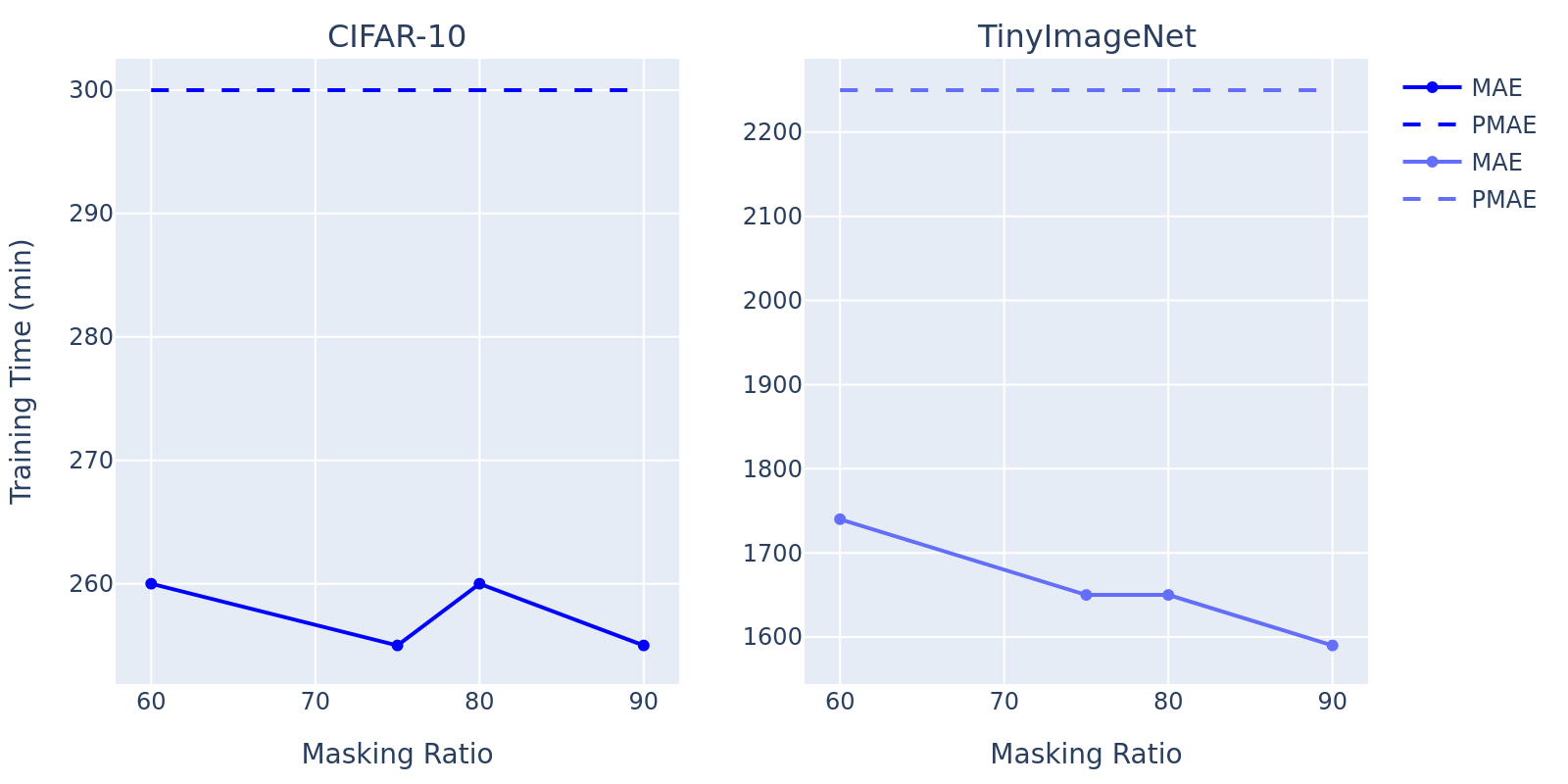}
    \caption{\textbf{Training time.} We report the training time in minutes for 800 training epochs using a ViT-T/8 architecture. For standard MAE we report numbers for various masking ratios.}
    \label{app_fig:ressources}
    \vspace{-10pt}  %
\end{figure}

All training runs were conducted on a single NVIDIA GeForce RTX 3090/NVIDIA GeForce RTX 4090/Quadro RTX 6000 GPUs or NVIDIA TITAN RTX each of which has 24GB of VRAM. \Cref{app_fig:ressources} reports the time taken for 800 training epochs using a ViT-T/8 architecture on a Quadro RTX 6000 GPU for CIFAR10 and TinyImageNet with both MAE and PMAE.

{\tiny
\begin{table}[h!]
    \centering
    \hspace{-30pt}
    \begin{minipage}{0.3\textwidth}
        \centering
        \vspace{-12pt}
         \begin{tabular}{ll}
            \toprule
            \textbf{config}               & \textbf{value} \\
            \midrule
            batch size                    & 512 \\
            base learning rate            & 0.00015 \\
            optimizer                     & AdamW \href{https://arxiv.org/abs/1711.05101}{[39]} \\
            betas (AdamW)                 & \small{$\beta_1, \beta_2 = 0.9, 0.95$} \\
            learning rate        & 0.0003 \\
            warmup steps                  & 40 \\
            weight decay                  & 0.05 \\
            \bottomrule
        \end{tabular}
        \caption{Training hyperparameters.}
        \label{app_tab:training}
    \end{minipage}
    \hspace{15pt}
    \begin{minipage}{0.3\textwidth}
        \centering
        \begin{tabular}{ll}
            \toprule
            \textbf{config}               & \textbf{value} \\
            \midrule
            batch size                    & 512 \\
            base learning rate            & 0.1 \\
            optimizer                     & SGD \href{https://arxiv.org/abs/2010.05374}{[6]} \\
            betas (SGD)                   & 0.9 \\
            learning rate                 & 0.2 \\
            warmup steps                  & 10 \\
            weight decay                  & 0 \\
            \bottomrule
        \end{tabular}
        \caption{Linear probing \\ hyperparameters.}
        \label{app_tab:evaluation_lin}
    \end{minipage}%
    \begin{minipage}{0.3\textwidth}
        \centering
        \begin{tabular}{ll}
            \toprule
            \textbf{config}               & \textbf{value} \\
            \midrule
            batch size                    & 512 \\
            base learning rate            & 0.001 \\
            optimizer                     & AdamW \href{https://arxiv.org/abs/1711.05101}{[39]} \\
            betas (AdamW)                 & \small{$\beta_1, \beta_2 = 0.9, 0.99$ } \\
            learning rate                 & 0.002 \\
            warmup steps                  & 5 \\
            weight decay                  & 0.5 \\
            \bottomrule
        \end{tabular}
        \caption{Fine-tuning \\ hyperparameters.}
        \label{app_tab:evaluation_fine}
    \end{minipage}
    \hspace{-30pt}
    \vspace{-0.5cm}
\end{table}}

\subsection{Additional Results}\label{app:results}
\subsubsection{Reconstruction of Masked Information}\label{app_sec:reconstruction}
\begin{figure}[ht]
    \vspace{-10pt}
    \centering
    \includegraphics[width=\linewidth]{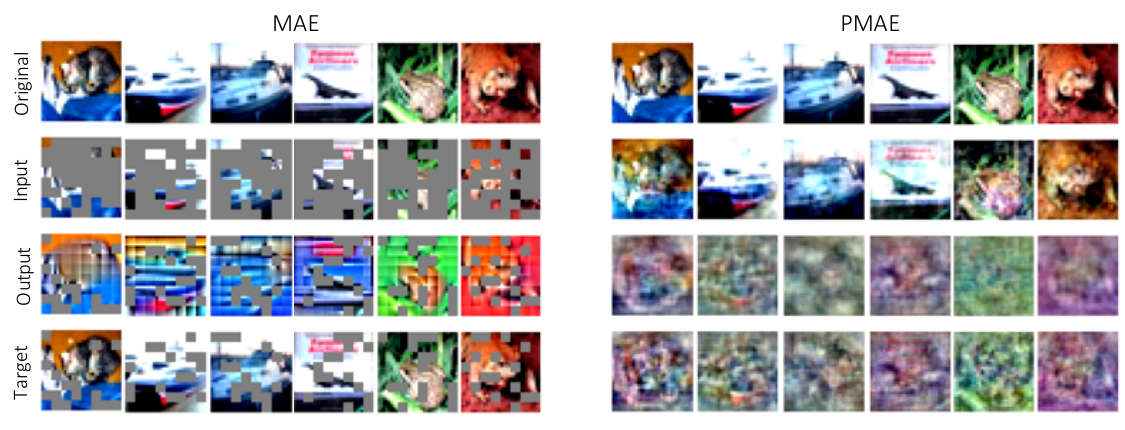}
    \vspace{-20pt}
    \caption{\small{\textbf{Reconstruction of masked information.} Example of original images, model input and output as well as the target for MAE (left) and PMAE (right) for CIFAR10. Training was performed for 1000 epochs with masking ratio of 75\% and 10\% for MAE and PMAE respectively. Reconstructions were obtained using the validation dataset.}}
    \vspace{-15pt}
    \label{fig:reconstructions}
\end{figure}

\Cref{fig:reconstructions} depicts the output of the decoder networks after 1000 training epochs for both MAE and PMAE. Interestingly, we observe that for PMAE, the model is able to well estimate the masked out principal components.
\subsubsection{Training Dynamics}
\begin{figure}[ht]
    \centering
    \vspace{-5pt}
    \begin{minipage}{0.35\textwidth}
        \begin{subfigure}%
        \centering
        \includegraphics[width=\textwidth]{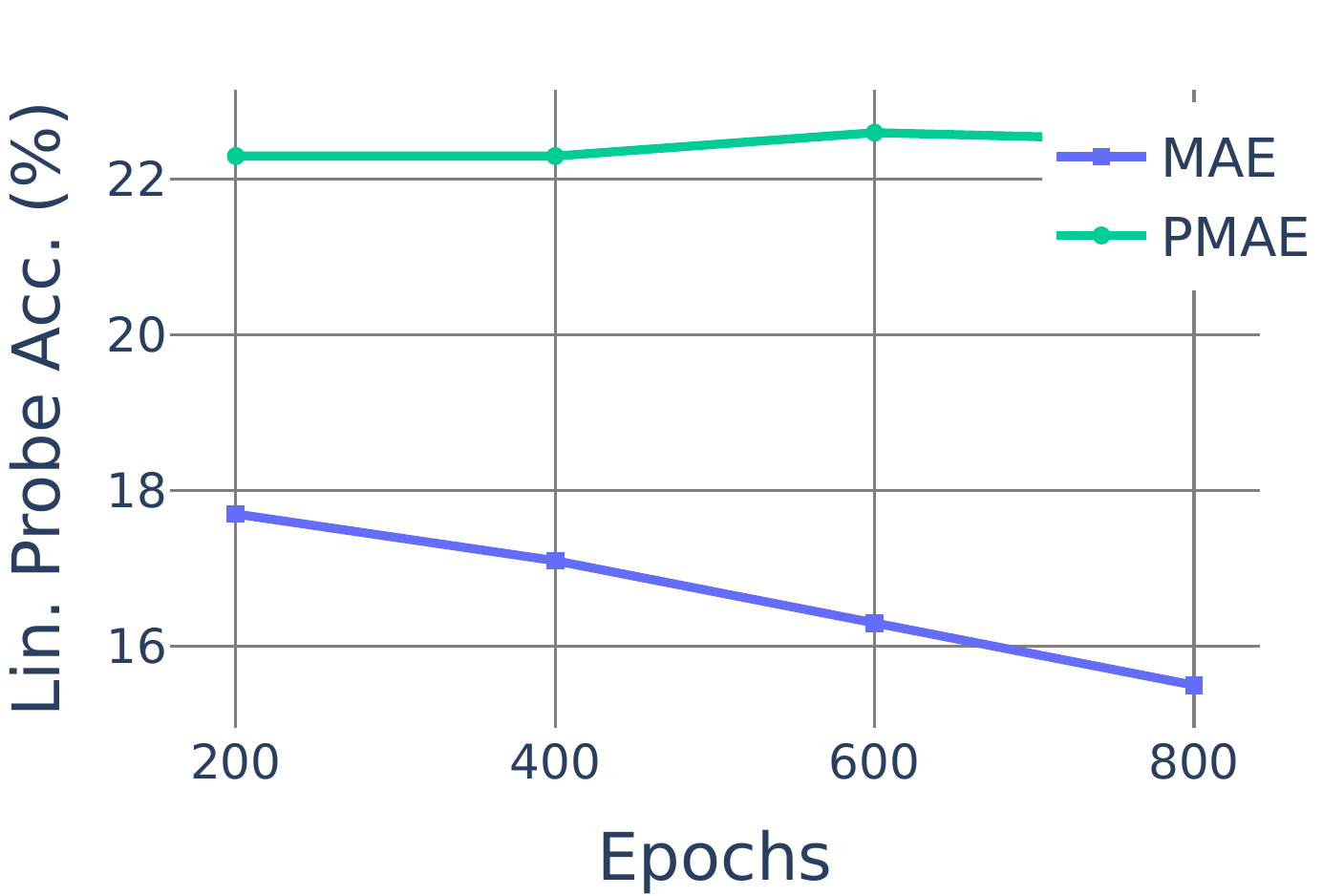}
        \vspace{-15pt}
        \caption{\small{TinyImageNet}}
        \label{fig:tiny_dataset}
    \end{subfigure}%
    \end{minipage}
    \begin{minipage}{0.35\textwidth}
        \begin{subfigure}
        \centering
        \includegraphics[width=\textwidth]{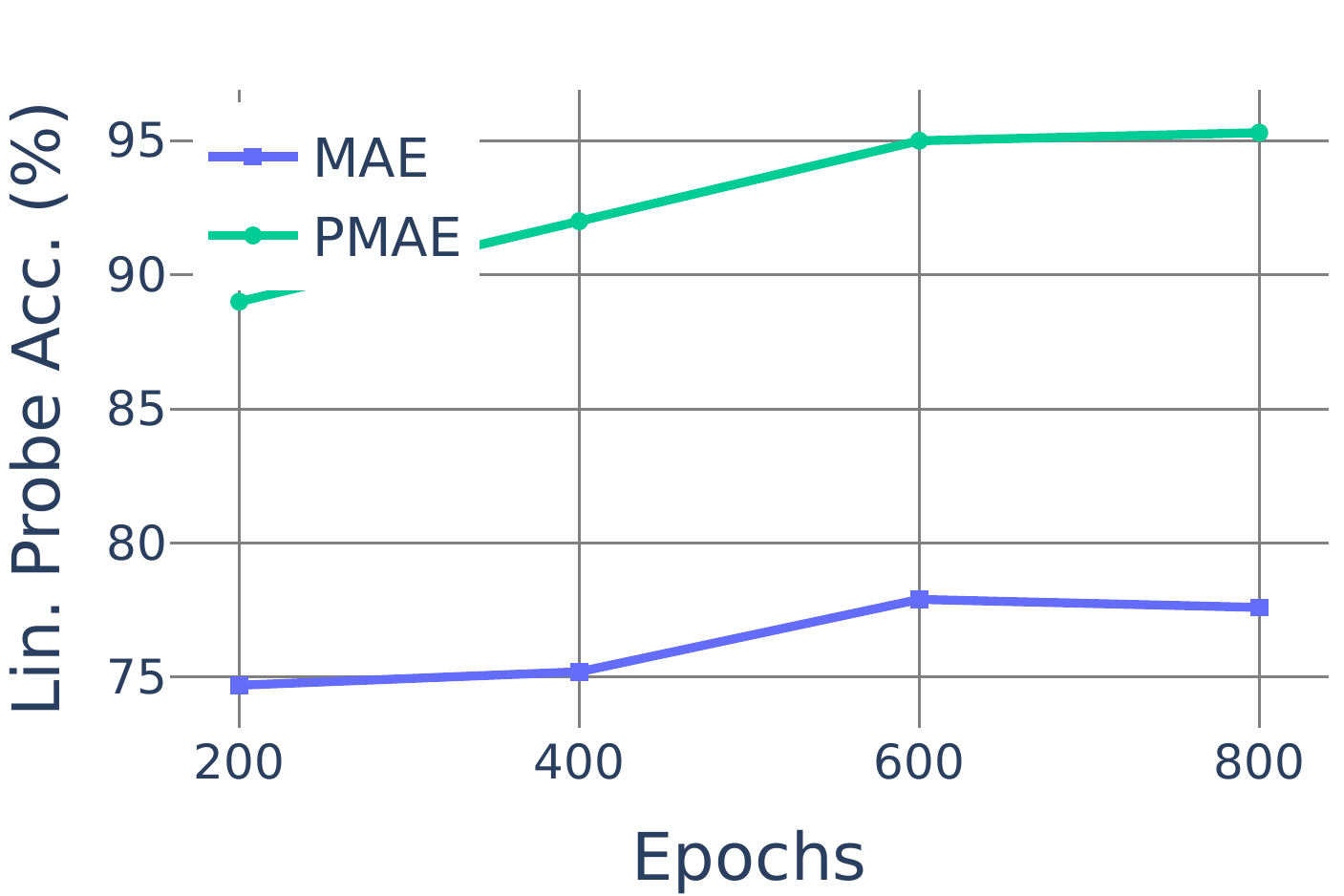}
        \vspace{-15pt}
        \caption{\small{BloodMNIST}}
        \label{fig:blood_dataset}
        \end{subfigure}
    \end{minipage}
    \begin{minipage}{0.35\textwidth}
        \begin{subfigure}%
        \centering
        \includegraphics[width=\textwidth]{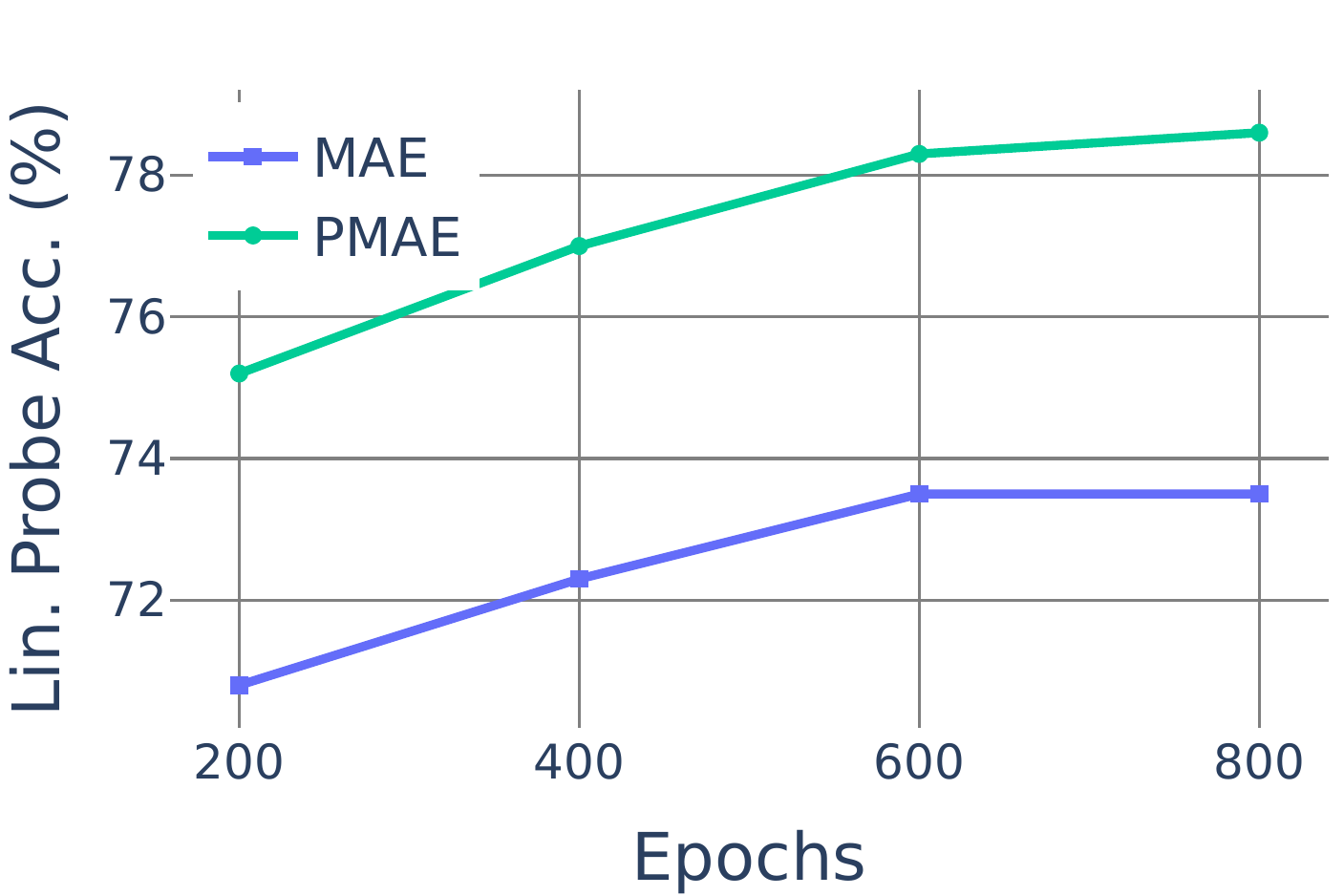}
        \vspace{-15pt}
        \caption{\small{DermaMNIST}}
        \label{fig:derma_dataset}
    \end{subfigure}%
    \end{minipage}
    \begin{minipage}{0.35\textwidth}
        \centering
        \includegraphics[width=\textwidth]{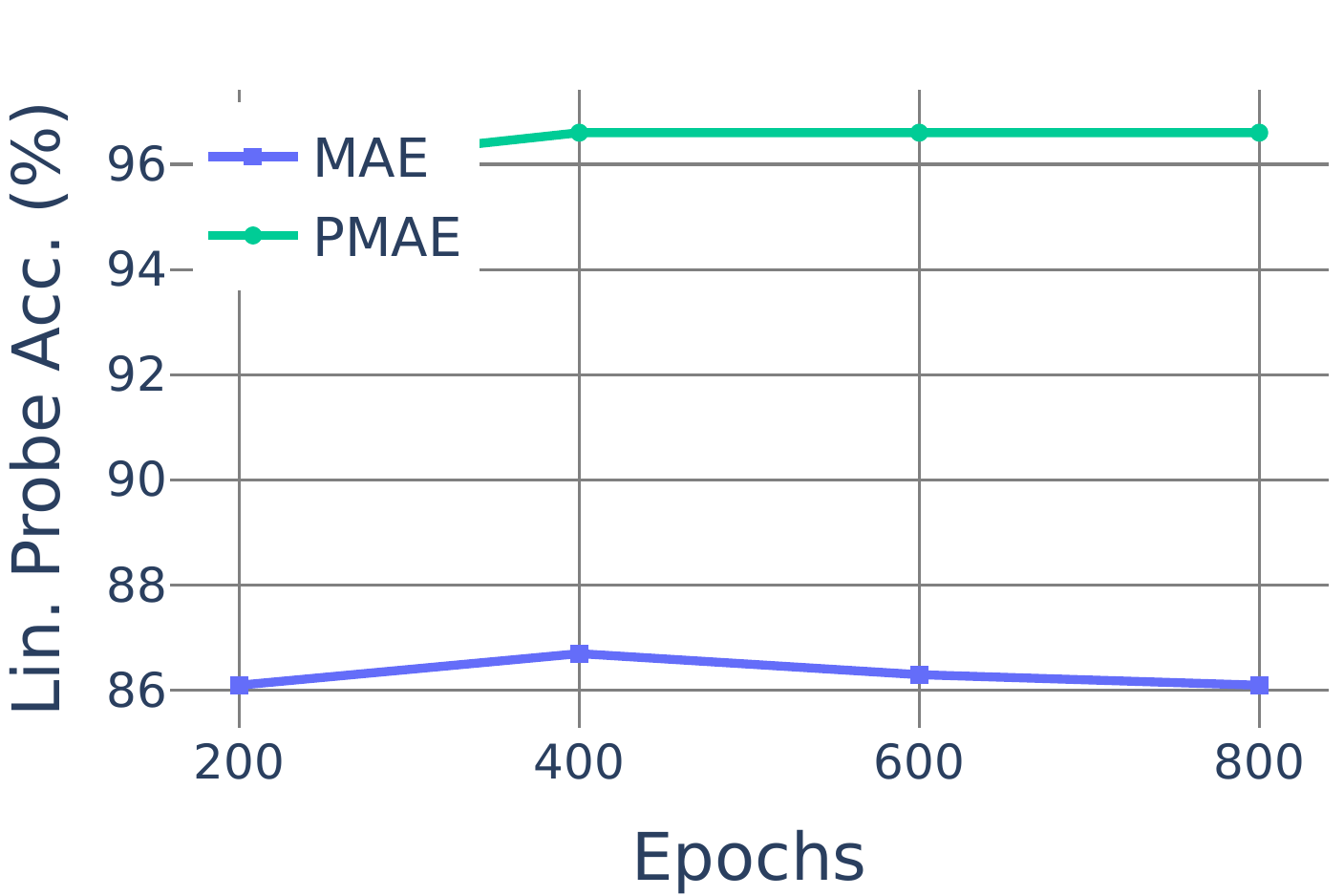}
        \vspace{-15pt}
        \caption{\small{PathMNIST}}
        \label{fig:path_dataset}
    \end{minipage}
    \caption{\small{\textbf{Performance Curves.} \looseness-1 Linear probe accuracy (\%) for TinyImageNet and MedMNIST datasets across training epochs. PMAE, after 200 epochs, outperforms MAE after 800 epochs.}}
    \label{app_fig:performance_curve}
\end{figure}

\Cref{app_fig:performance_curve} displays the linear probe accuracy for varying training epoch checkpoints. Similar to \Cref{fig:mae_hyperparameters} (right) we observe that PMAE after 200 epochs outperforms MAE after 800 epochs.

\subsubsection{$k$-Nearest Neighbor Evaluation}
\looseness-1 \Cref{app_tab:results} shows the classification accuracy for CIFAR10, TinyImageNet, and MedMNIST datasets using a $k$-NN classifier in place of a linear or MLP probe as presented in \Cref{sec:results} and \Cref{app:reconstruct_pc}.  

\begin{table*}[ht]
\centering
\caption{\small{\textbf{Evaluation with $\bm{k}$-NN.} $k$-Nearest Neighbors top-$1\%$ accuracy for CIFAR10, TinyImageNet and MedMNIST for masking in pixel (MAE) and principal component (PMAE) space with the standard $75\%$ masking ratio (std), oracles (ocl) and randomized (rd) masking ratios. $\star$ and $\star\star$ refer to representations trained with \Cref{eq:pmae_lossA} and \Cref{eq:pmae} respectively.}}
\label{app_tab:results}
\resizebox{0.7\columnwidth}{!}{
\begin{tabular}{clccccc}
\toprule
&& \small{\textbf{CIFAR10}}   & \small{\textbf{TinyImageNet}}  & \small{\textbf{DermaMNIST}}   & \small{\textbf{BloodMNIST}} & \small{\textbf{PathMNIST}}  \\
\midrule
\multirow{7}{*}{\textbf{$\bm k$-NN}} 
& \cellcolor{gray!10} MAE$_{\text{std}}$  & \cellcolor{gray!10} $38.3$  & \cellcolor{gray!10} $10.0$  & \cellcolor{gray!10} $71.1$  & \cellcolor{gray!10} $65.7$  & \cellcolor{gray!10} $92.1$  \\
& \cellcolor{gray!20} MAE$_{\text{ocl}}$  & \cellcolor{gray!20} $47.6$  & \cellcolor{gray!20} $12.5$  & \cellcolor{gray!20} $69.9$  & \cellcolor{gray!20} $73.6$  & \cellcolor{gray!20} $94.6$  \\
& \cellcolor{gray!20} PMAE$_{\text{ocl}}^{\star}$ & \cellcolor{gray!20} $\mathbf{55.3}$  & \cellcolor{gray!20} $\mathbf{14.2}$  & \cellcolor{gray!20} $\mathbf{76.6}$  & \cellcolor{gray!20} $\mathbf{93.1}$  & \cellcolor{gray!20} $\mathbf{99.6}$  \\
& \cellcolor{gray!20} PMAE$_{\text{ocl}}^{\star\star}$ & \cellcolor{gray!20} $48.1$  & \cellcolor{gray!20} $9.6$  & \cellcolor{gray!20} $74.7$  & \cellcolor{gray!20} $84.5$  & \cellcolor{gray!20} $99.1$  \\
& MAE$_{\text{rd}}$   & $40.3$  & $7.6$   & $71.6$  & $82.7$  & $\mathbf{96.0}$  \\
& PMAE$_{\text{rd}}^{\star}$  & $41.8$  & $\mathbf{11.2}$  & $\mathbf{72.1}$  & $\mathbf{83.6}$  & $95.5$  \\
& PMAE$_{\text{rd}}^{\star\star}$  & $\mathbf{49.6}$  & $9.5$  & $70.6$  & $76.0$  & $94.8$  \\
\bottomrule
\end{tabular}
}
\end{table*}

\subsubsection{Reconstructing in pixel vs. principal component space}\label{app:reconstruct_pc}
\begin{figure*}[ht]
    \centering
    \includegraphics[width=0.8\linewidth]{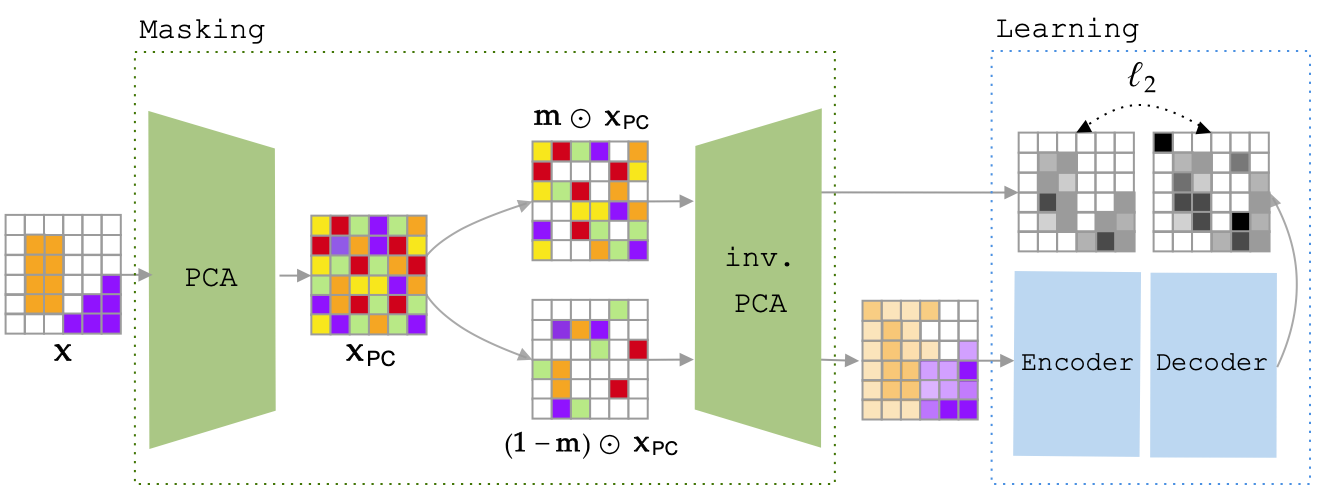}
    \caption{\small{\looseness-1 \textbf{Overview of the Principal Masked Autoencoder (PMAE) with reconstruction targets in the observation space.} A Principal Masked Autoencoder (PMAE) differs from a vanilla MAE by performing the masking in the space of principal components $\xb_\text{PC}\!=\!\text{PCA}(\xb)$ rather than in the pixel space. The masked and visible principal component $\mb \odot \xb_{\text{PC}}$ and $(\bm 1\!-\!\mb) \odot \xb_{\text{PC}}$  are then projected back into the space of pixels and serve as the %
    reconstruction target and input for an encoder-decoder architecture, respectively. 
    }}
    \label{fig:overview_A}
    \vspace{-10pt}
\end{figure*}

We further investigate the impact of the domain (i.e., pixel vs. pc space) in which the reconstruction error is minimized on downstream performance. In \Cref{fig:overview_A}, we present an alternative to \Cref{fig:overview} in which the training objective receives a set of pixels in place of principal components. In \Cref{fig:overview_A}, the masked out principal components are projected back into pixel space and the decoder's output is kept as is. The training objective (\Cref{eq:pmae}) then minimizes the Euclidean distance between the ground truth masked principal components projected back to pixel space and the decoder's output. Instead in \Cref{fig:overview} and \Cref{eq:pmae_lossA}, the training objective minimizes the Euclidean distance between the ground truth and the predicted masked principal components. \Cref{eq:pmae} becomes a modified version of \Cref{eq:pmae_lossA}:

\begin{equation}\label{eq:pmae}
    \mathcal{L}_{\text{PMAE}}(\mathbf{x},\mathbf{m};\bm\theta,\bm\phi) 
    = \left\| \left(g_{\bm{\theta}}\circ f_{\bm{\phi}}\circ t^{-1}\left(\mathbf{m}\odot t\left(\mathbf{x}\right)\right)\right)
    - t^{-1}\left(\left(\mathbf{1}-\mathbf{m}\right)\odot t(\mathbf{x})\right) \right\|_2^2,
\end{equation}

\Cref{tab:results_A} presents the downstream image classification performance achieved when training representations with \Cref{eq:pmae}. In particular, we report results obtained using a linear and MLP probe in the standard, oracle and randomized masking settings. Despite lower performance gains as the ones presented with \Cref{eq:pmae_lossA} in \Cref{tab:results}, PMAE consistently outperforms MAE across all five datasets, demonstrating substantial improvements. In \Cref{tab:results} we report an average performance gain of 9.6 percentage points across datasets over the MAE baseline, while \Cref{tab:results_A} reports an average performance gain of 6.6 percentage points. These findings also support our claims that the space of principal components constitutes a meaningful masking space for Masking Image Modelling learning paradigms.

\subsubsection{Masking Ratio Ablation}
\begin{figure}
    \centering
    \begin{subfigure}
        \centering
        \includegraphics[width=0.9\linewidth]{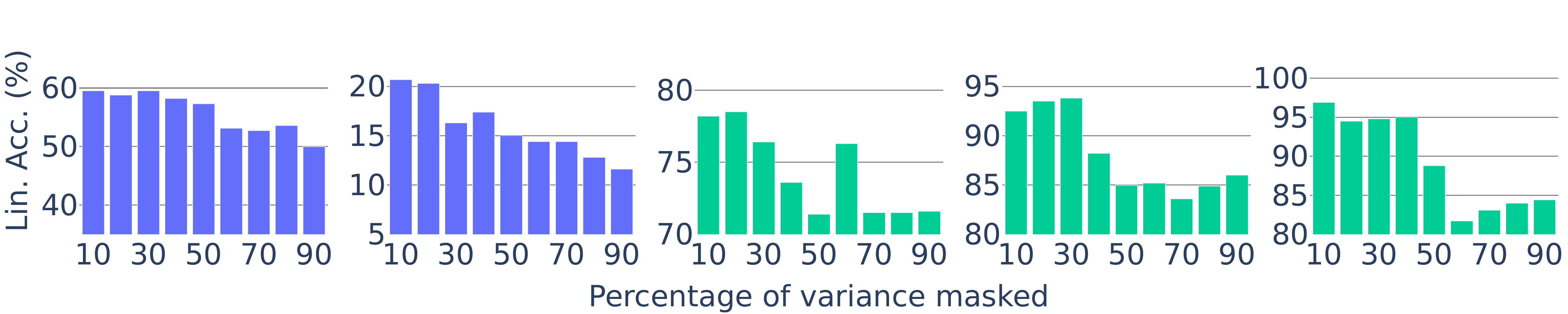}
        \label{fig:pmae_masking_ratio}
    \end{subfigure}
    \begin{subfigure}
        \centering
        \includegraphics[width=0.9\linewidth]{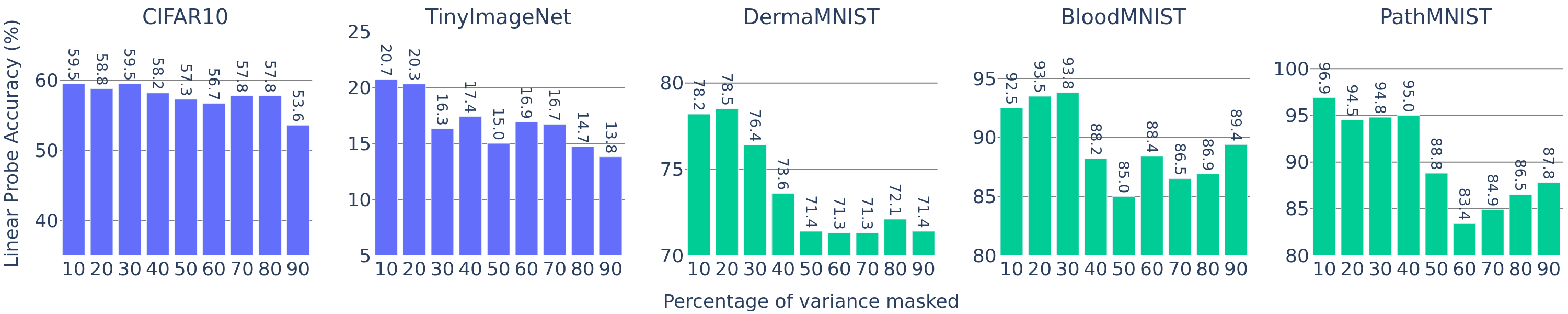}
        \label{fig:pmae_masking_ratio_old}
    \end{subfigure}
    \vspace{-15pt}
    \caption{\small{\textbf{Impact of the Masking Ratio.} (\textit{top}) PMAE trained with \Cref{eq:pmae} linear probing accuracy for varying masking ratios. We observe a close-to-optimal performance across datasets for $10$ to $20\%$ of the variance masked. \textit{(bottom)} PMAE trained with \Cref{eq:pmae} linear probing accuracy \textit{without} data augmentations at evaluation. We observe a significant performance improvement without data augmentations.}}
\label{fig:pmae_masking_ratio_combined}
\end{figure}

\looseness-1  In \Cref{fig:pmae_masking_ratio}, we perform an ablation across the masking ratio for representations learned with \Cref{eq:pmae}. We draw similar conclusions to the ones drawn from \Cref{fig:pmae_masking_ratio}: the optimal masking ratio across datasets lies between $10$ and $20\%$ of the variance masked. Above these ratios, we observe a performance drop across datasets. \Cref{fig:pmae_masking_ratio_old} also shows the impact of the use of data augmentations during training of the linear probe on the downstream accuracy. We see a significant increase in performance for PMAE when dropping data augmentations for the training of the linear probe but still resort to keeping these augmentations for the main results presented in \Cref{tab:results} to ensure a fair comparison with the MAE.

\subsubsection{Beyond PCA}\label{app:kernel}
Our work shows evidence that PCA offers a meaningful masking space. In \Cref{sec:explanation}, we motivate our choice by observing that principal components capture global rather than local features of an image. In this section, we go beyond PCA and explore non-linear matrix factorization methods as a proof of concept for future research. In particular, we explore Kernel PCA \citep{scholkopf1997kernel} with a Radial Basis Function (RBF) kernel with the kernel coefficient set to $3.10^{-4}$. In kernel PCA, the spectral decomposition is performed not on the data itself but rather on a modified version of it: the standardized data is mapped to a high-dimensional space via a non-linear kernel function.

\begin{wraptable}{r}{0.5\textwidth}  %
\centering
\vspace{-10pt}
\caption{\small{Linear and MLP probe accuracy for CIFAR10 for random masking in pixel (MAE), in principal component space (PMAE), and in Kernel PCA space (KMAE) with the standard $75\%$ masking ratio (std) and oracles (ocl). $^*$ refers to ours.}}
\label{tab:results_kernel}
\resizebox{0.48\textwidth}{!}{
\begin{tabular}{ccccc}
\toprule
& \small{\textbf{MAE}$_{\text{std}}$} & \small{\textbf{MAE}$_{\text{ocl}}$}   & \small{\textbf{PMAE}$_{\text{ocl}}^*$}  & \small{\textbf{KMAE}$_{\text{ocl}}^*$}  \\
\midrule
\small{Linear} & $41.7$  & $50.7$  & $59.0$ &  $\bm{64.1}$\\
\small{MLP} & $34.0$  & $55.2$  & $64.1$ & $\bm{68.6}$\\
\midrule
\end{tabular}
}
\end{wraptable}

In \Cref{tab:results_kernel}, we present results on the CIFAR10 dataset and show the image classification accuracy using a linear and MLP probe. We compare a vanilla MAE with our PMAE and KMAE which relies on Kernel PCA for optimal masking ratios (30\% of the variance is masked out). For KMAE, we use the setting presented in \Cref{app:reconstruct_pc} and use \Cref{eq:pmae_lossA} as the training objective.

The results reveal a significant performance improvement when employing a non-linear image transformation. KMAE achieves an average gain of $13.3$ and $5$ percentage points compared to the standard Masked Autoencoder \citep{he_masked_2021} and PMAE, respectively. Although these findings are preliminary and based on a single mid-scale dataset, they highlight the potential of non-linear transformations and further emphasize the value of spectral decomposition as a meaningful for masked image modeling paradigms.

\begin{table}
\centering
\caption{\small{Linear \& MLP probe and fine-tuning top-$1\%$ accuracy for CIFAR10, TinyImageNet and MedMNIST datasets for random masking in pixel (MAE) and principal component (PMAE) space with reconstruction target in pixel space with the standard $75\%$ masking ratio (std), oracles (ocl) and randomized masking ratios (rd). Representations are learnt from \Cref{eq:pmae}. $^*$ refers to ours.}}
\resizebox{0.7\textwidth}{!}{
\begin{tabular}{clccccc}
\toprule
&& \small{\textbf{CIFAR10}}   & \small{\textbf{TinyImageNet}}  & \small{\textbf{DermaMNIST}}   & \small{\textbf{BloodMNIST}} & \small{\textbf{PathMNIST}}  \\
\midrule
\multirow{5}{*}{\textbf{Linear}} 
& \cellcolor{gray!25} MAE$_{\text{std}}$  & \cellcolor{gray!25} $41.7$  & \cellcolor{gray!25} $11.5$  & \cellcolor{gray!25} $72.4$  & \cellcolor{gray!25} $73.4$  & \cellcolor{gray!25} $83.4$  \\
& \cellcolor{gray!10} MAE$_{\text{ocl}}$  & \cellcolor{gray!10} $50.7$  & \cellcolor{gray!10} $15.5$  & \cellcolor{gray!10} $73.7$  & \cellcolor{gray!10} $78.6$  & \cellcolor{gray!10} $86.4$  \\
& \cellcolor{gray!10} PMAE$_{\text{ocl}}^{*}$ & \cellcolor{gray!10} $\mathbf{55.1}$  & \cellcolor{gray!10} $\mathbf{17.4}$  & \cellcolor{gray!10} $\mathbf{77.4}$  & \cellcolor{gray!10} $\mathbf{91.0}$  & \cellcolor{gray!10} $\mathbf{97.0}$  \\
& MAE$_{\text{rd}}$   & $41.9$  & $7.5$   & $72.4$  & $83.2$  & $85.6$  \\
& PMAE$_{\text{rd}}^{*}$  & $\mathbf{56.0}$  & $\mathbf{15.1}$  & $\mathbf{74.5}$  & $\mathbf{85.9}$  & $\mathbf{87.5}$  \\
\midrule
\multirow{5}{*}{\textbf{MLP}} 
& \cellcolor{gray!25} MAE$_{\text{std}}$  & \cellcolor{gray!25} $34.0$  & \cellcolor{gray!25} $15.5$  & \cellcolor{gray!25} $72.2$  & \cellcolor{gray!25} $68.6$  & \cellcolor{gray!25} $92.6$  \\
& \cellcolor{gray!10} MAE$_{\text{ocl}}$  & \cellcolor{gray!10} $55.2$  & \cellcolor{gray!10} $\mathbf{22.2}$  & \cellcolor{gray!10} $74.4$  & \cellcolor{gray!10} $75.8$  & \cellcolor{gray!10} $95.1$  \\
& \cellcolor{gray!10} PMAE$_{\text{ocl}}^{*}$ & \cellcolor{gray!10} $\mathbf{61.5}$  & \cellcolor{gray!10} $22.1$  & \cellcolor{gray!10} $\mathbf{79.6}$  & \cellcolor{gray!10} $\mathbf{91.0}$  & \cellcolor{gray!10} $\mathbf{98.8}$  \\
& MAE$_{\text{rd}}$   & $38.5$  & $11.6$   & $66.9$  & $70.6$  & $95.7$  \\
& PMAE$_{\text{rd}}^{*}$  & $\mathbf{62.2}$  & $\mathbf{19.5}$  & $\mathbf{75.3}$  & $\mathbf{84.4}$  & $\mathbf{97.0}$  \\
\bottomrule
\end{tabular}
}
\label{tab:results_A}
\end{table}

\end{document}